\documentclass{article}

\usepackage{titling}
\usepackage{standalone}
\usepackage[utf8]{inputenc}
\usepackage[T1]{fontenc}
\usepackage{lmodern}
\usepackage{xr}
\usepackage{xr-hyper}
\usepackage[hidelinks]{hyperref}
\usepackage{url}
\usepackage{booktabs}
\usepackage{amsfonts}
\usepackage{nicefrac}
\usepackage{microtype}
\usepackage{cleveref}
\usepackage{lipsum}
\usepackage{graphicx}
\usepackage{doi}
\usepackage{siunitx}
\usepackage{pdflscape}
\usepackage{caption}
\usepackage{subcaption}
\usepackage{authblk}
\usepackage{threeparttable}

\usepackage[
	style=numeric-comp,
	backend=biber,
	sorting=none,
	terseinits=true,
	giveninits=true,
	minnames=6,
	maxnames=6,
	url=false,
	doi=true,
	isbn=false,
	eprint=false
]{biblatex}

\DeclareNameAlias{default}{family-given}

\DeclareFieldFormat{labelnumberwidth}{[#1]}
\renewbibmacro{in:}{}

\DeclareFieldFormat[article]{title}{#1}
\DeclareFieldFormat[article]{journaltitle}{#1}
\DeclareFieldFormat[article]{volume}{#1}
\DeclareFieldFormat[article]{number}{(#1)}
\DeclareFieldFormat[article]{pages}{:#1}
\DeclareFieldFormat{doi}{\mkbibacro{DOI}\addcolon\space\url{https://doi.org/#1}}
\DeclareFieldFormat[online]{url}{Available from: \url{#1}}
\DeclareFieldFormat[online]{title}{#1}
\DeclareFieldFormat[online]{version}{Version #1}

\newcommand*{\shortmonth}[1]{%
	\ifcase#1\relax
	\or Jan\or Feb\or Mar\or Apr\or May\or Jun%
	\or Jul\or Aug\or Sep\or Oct\or Nov\or Dec\fi}

\DeclareBibliographyDriver{article}{%
	\usebibmacro{bibindex}%
	\usebibmacro{begentry}%
	\printnames{author}%
	\setunit{\addperiod\space}%
	\printfield{title}%
	\setunit{\addperiod\space}%
	\iffieldundef{shortjournal}{%
		\printfield{journaltitle}%
	}{%
		\printfield{shortjournal}%
	}%
	\setunit{\addperiod\space}%
	\printfield{year}%
	\setunit{\addsemicolon}%
	\printfield{volume}%
	\printfield{number}%
	\printfield{pages}%
	\setunit{\addperiod\space}%
	\printfield{doi}%
	\setunit{\addperiod}%
	\usebibmacro{finentry}}

\DeclareBibliographyDriver{online}{%
	\usebibmacro{bibindex}%
	\usebibmacro{begentry}%
	\printnames{author}%
	\setunit{\addperiod\space}%
	\printfield{title}%
	\setunit{\addperiod\space}%
	\printlist{publisher}%
	\setunit{\space}%
	\printtext{[Preprint]}%
	\setunit{\space}%
	\printfield{year}%
	\setunit{\addperiod\space}%
	\printfield{url}%
	\setunit{\addperiod}%
	\usebibmacro{finentry}}

\usepackage{arxiv}

\crefname{figure}{}{}
\Crefname{figure}{}{}
\crefname{table}{}{}
\Crefname{table}{}{}
\crefname{section}{}{}
\Crefname{section}{}{}
\crefname{equation}{}{}
\Crefname{equation}{}{}
\crefname{appendix}{}{}
\Crefname{appendix}{}{}

\newcommand{\tableOne}{\begin{tabular}{@{}lcccccccc@{}}%
\toprule%
\textbf{Cohort}&\multicolumn{1}{c}{\textbf{STG}}&\multicolumn{2}{c}{\textbf{STHLM3}}&\multicolumn{2}{c}{\textbf{SUH}}&\multicolumn{1}{c}{\textbf{AMU}}&\multicolumn{1}{c}{\textbf{MUL}}&\multicolumn{1}{c}{\textbf{SCH}}\\%
\cmidrule(lr){2-2}%
\cmidrule(lr){3-4}%
\cmidrule(lr){5-6}%
\cmidrule(lr){7-7}%
\cmidrule(lr){8-8}%
\cmidrule(lr){9-9}%
\textbf{Split}&\textbf{Train}&\textbf{Train}&\textbf{Test}&\textbf{Train}&\textbf{Test}&\textbf{Test}&\textbf{Test}&\textbf{Test}\\%
\midrule%
\midrule%
\multicolumn{9}{c}{\textbf{Patients}}\\%
\midrule%
\textbf{\textit{n}}&67&287&140&76&31&43&49&12\\%
\multicolumn{9}{l}{\hspace{-3mm}\textbf{Age, years}}\\%
\hspace{3mm}$\leq$49&0 (0\%)&0 (0\%)&1 ($<$1\%)&0 (0\%)&0 (0\%)&0 (0\%)&1 (2\%)&0 (0\%)\\%
\hspace{3mm}50-54&1 (2\%)&17 (6\%)&6 (4\%)&3 (4\%)&1 (3\%)&0 (0\%)&1 (2\%)&0 (0\%)\\%
\hspace{3mm}55-59&2 (4\%)&31 (11\%)&18 (13\%)&3 (4\%)&2 (6\%)&0 (0\%)&1 (2\%)&3 (25\%)\\%
\hspace{3mm}60-64&4 (9\%)&80 (28\%)&35 (25\%)&14 (18\%)&1 (3\%)&0 (0\%)&6 (12\%)&3 (25\%)\\%
\hspace{3mm}65-69&4 (9\%)&149 (52\%)&72 (51\%)&14 (18\%)&6 (19\%)&0 (0\%)&9 (18\%)&3 (25\%)\\%
\hspace{3mm}$\geq$70&36 (77\%)&10 (3\%)&8 (6\%)&42 (55\%)&21 (68\%)&0 (0\%)&31 (63\%)&3 (25\%)\\%
\hspace{3mm}Missing&20&0&0&0&0&43&0&0\\%
\multicolumn{9}{l}{\hspace{-3mm}\textbf{PSA, ng/mL}}\\%
\hspace{3mm}$<$3&3 (7\%)&45 (16\%)&16 (11\%)&4 (5\%)&0 (0\%)&1 (2\%)&0 (0\%)&0 (0\%)\\%
\hspace{3mm}3-$<$5&0 (0\%)&88 (31\%)&56 (40\%)&5 (7\%)&1 (3\%)&1 (2\%)&0 (0\%)&1 (11\%)\\%
\hspace{3mm}5-$<$10&7 (16\%)&76 (26\%)&39 (28\%)&34 (45\%)&13 (42\%)&11 (26\%)&0 (0\%)&4 (44\%)\\%
\hspace{3mm}$\geq$10&35 (78\%)&78 (27\%)&29 (21\%)&32 (43\%)&17 (55\%)&30 (70\%)&0 (0\%)&4 (44\%)\\%
\hspace{3mm}Missing&22&0&0&1&0&0&49&3\\%
\midrule%
\multicolumn{9}{c}{\textbf{Whole Slide Images}}\\%
\midrule%
\textbf{\textit{n\tnote{*}}}&79&1,051&608&150&50&73&137&56\\%
\textbf{Physical slides}&79&411&211&150&50&73&137&56\\%
\textbf{Cribriform}&27 (34\%)&81 (20\%)&43 (20\%)&47 (31\%)&19 (38\%)&28 (38\%)&55 (40\%)&11 (20\%)\\%
\multicolumn{9}{l}{\hspace{-2mm}\textbf{Gleason score}}\\%
\hspace{3mm}3 + 3&0 (0\%)&0 (0\%)&0 (0\%)&0 (0\%)&0 (0\%)&0 (0\%)&0 (0\%)&3 (5\%)\\%
\hspace{3mm}3 + 4&0 (0\%)&61 (15\%)&25 (12\%)&68 (45\%)&13 (26\%)&0 (0\%)&21 (15\%)&18 (32\%)\\%
\hspace{3mm}3 + 5&1 (1\%)&11 (3\%)&1 ($<$1\%)&0 (0\%)&0 (0\%)&0 (0\%)&0 (0\%)&5 (9\%)\\%
\hspace{3mm}4 + 3&2 (3\%)&131 (32\%)&74 (35\%)&37 (25\%)&16 (32\%)&0 (0\%)&38 (28\%)&19 (34\%)\\%
\hspace{3mm}4 + 4&17 (22\%)&158 (38\%)&73 (35\%)&25 (17\%)&15 (30\%)&0 (0\%)&35 (26\%)&5 (9\%)\\%
\hspace{3mm}4 + 5&27 (34\%)&39 (9\%)&34 (16\%)&18 (12\%)&4 (8\%)&0 (0\%)&28 (20\%)&6 (11\%)\\%
\hspace{3mm}5 + 3&0 (0\%)&1 ($<$1\%)&0 (0\%)&0 (0\%)&0 (0\%)&0 (0\%)&0 (0\%)&0 (0\%)\\%
\hspace{3mm}5 + 4&19 (24\%)&6 (1\%)&2 ($<$1\%)&2 (1\%)&2 (4\%)&0 (0\%)&15 (11\%)&0 (0\%)\\%
\hspace{3mm}5 + 5&13 (16\%)&4 ($<$1\%)&2 ($<$1\%)&0 (0\%)&0 (0\%)&0 (0\%)&0 (0\%)&0 (0\%)\\%
\hspace{3mm}Missing&0&0&0&0&0&73&0&0\\%
\multicolumn{9}{l}{\hspace{-2mm}\textbf{ISUP}}\\%
\hspace{3mm}1&0 (0\%)&0 (0\%)&0 (0\%)&0 (0\%)&0 (0\%)&0 (0\%)&0 (0\%)&3 (5\%)\\%
\hspace{3mm}2&0 (0\%)&61 (15\%)&25 (12\%)&68 (45\%)&13 (26\%)&0 (0\%)&21 (15\%)&18 (32\%)\\%
\hspace{3mm}3&2 (3\%)&131 (32\%)&74 (35\%)&37 (25\%)&16 (32\%)&0 (0\%)&38 (28\%)&19 (34\%)\\%
\hspace{3mm}4&18 (23\%)&170 (41\%)&74 (35\%)&25 (17\%)&15 (30\%)&0 (0\%)&35 (26\%)&10 (18\%)\\%
\hspace{3mm}5&59 (75\%)&49 (12\%)&38 (18\%)&20 (13\%)&6 (12\%)&0 (0\%)&43 (31\%)&6 (11\%)\\%
\hspace{3mm}Missing&0&0&0&0&0&73&0&0\\\bottomrule%
\end{tabular}}
\newcommand{\tableOneSplit}{\begin{tabular}{@{}lcccc@{}}%
\toprule%
\textbf{Split}&\textbf{Train}&\textbf{Internal test}&\textbf{External test}&\textbf{Overall}\\%
\midrule%
\midrule%
\multicolumn{5}{c}{\textbf{Patients}}\\%
\midrule%
\textbf{\textit{n}}&430&171&104&705\\%
\multicolumn{5}{l}{\hspace{-3mm}\textbf{Age, years}}\\%
\hspace{3mm}$\leq$49&0 (0\%)&1 ($<$1\%)&1 (2\%)&2 ($<$1\%)\\%
\hspace{3mm}50-54&21 (5\%)&7 (4\%)&1 (2\%)&29 (5\%)\\%
\hspace{3mm}55-59&36 (9\%)&20 (12\%)&4 (7\%)&60 (9\%)\\%
\hspace{3mm}60-64&98 (24\%)&36 (21\%)&9 (15\%)&143 (22\%)\\%
\hspace{3mm}65-69&167 (41\%)&78 (46\%)&12 (20\%)&257 (40\%)\\%
\hspace{3mm}$\geq$70&88 (21\%)&29 (17\%)&34 (56\%)&151 (24\%)\\%
\hspace{3mm}Missing&20&0&43&63\\%
\multicolumn{5}{l}{\hspace{-3mm}\textbf{PSA, ng/mL}}\\%
\hspace{3mm}$<$3&52 (13\%)&16 (9\%)&1 (2\%)&69 (11\%)\\%
\hspace{3mm}3-$<$5&93 (23\%)&57 (33\%)&2 (4\%)&152 (24\%)\\%
\hspace{3mm}5-$<$10&117 (29\%)&52 (30\%)&15 (29\%)&184 (29\%)\\%
\hspace{3mm}$\geq$10&145 (36\%)&46 (27\%)&34 (65\%)&225 (36\%)\\%
\hspace{3mm}Missing&23&0&52&75\\%
\midrule%
\multicolumn{5}{c}{\textbf{Whole Slide Images}}\\%
\midrule%
\textbf{\textit{n\tnote{*}}}&1,280&658&266&2,204\\%
\textbf{Physical slides}&640&261&266&1,167\\%
\textbf{Cribriform}&155 (24\%)&62 (24\%)&94 (35\%)&311 (27\%)\\%
\multicolumn{5}{l}{\hspace{-2mm}\textbf{Gleason score}}\\%
\hspace{3mm}3 + 3&0 (0\%)&0 (0\%)&3 (2\%)&3 ($<$1\%)\\%
\hspace{3mm}3 + 4&129 (20\%)&38 (15\%)&39 (20\%)&206 (19\%)\\%
\hspace{3mm}3 + 5&12 (2\%)&1 ($<$1\%)&5 (3\%)&18 (2\%)\\%
\hspace{3mm}4 + 3&170 (27\%)&90 (34\%)&57 (30\%)&317 (29\%)\\%
\hspace{3mm}4 + 4&200 (31\%)&88 (34\%)&40 (21\%)&328 (30\%)\\%
\hspace{3mm}4 + 5&84 (13\%)&38 (15\%)&34 (18\%)&156 (14\%)\\%
\hspace{3mm}5 + 3&1 ($<$1\%)&0 (0\%)&0 (0\%)&1 ($<$1\%)\\%
\hspace{3mm}5 + 4&27 (4\%)&4 (2\%)&15 (8\%)&46 (4\%)\\%
\hspace{3mm}5 + 5&17 (3\%)&2 ($<$1\%)&0 (0\%)&19 (2\%)\\%
\hspace{3mm}Missing&0&0&73&73\\%
\multicolumn{5}{l}{\hspace{-2mm}\textbf{ISUP}}\\%
\hspace{3mm}1&0 (0\%)&0 (0\%)&3 (2\%)&3 ($<$1\%)\\%
\hspace{3mm}2&129 (20\%)&38 (15\%)&39 (20\%)&206 (19\%)\\%
\hspace{3mm}3&170 (27\%)&90 (34\%)&57 (30\%)&317 (29\%)\\%
\hspace{3mm}4&213 (33\%)&89 (34\%)&45 (23\%)&347 (32\%)\\%
\hspace{3mm}5&128 (20\%)&44 (17\%)&49 (25\%)&221 (20\%)\\%
\hspace{3mm}Missing&0&0&73&73\\\bottomrule%
\end{tabular}}
\newcommand{\tableOneCribriform}{\begin{tabular}{@{}lcc@{}}%
\toprule%
\textbf{}&\textbf{Cribriform}&\textbf{Non{-}cribriform}\\%
\midrule%
\midrule%
\multicolumn{3}{c}{\textbf{Patients}}\\%
\midrule%
\textbf{\textit{n}}&221&587\\%
\multicolumn{3}{l}{\hspace{-3mm}\textbf{Age, years}}\\%
\hspace{3mm}$\leq$49&1 ($<$1\%)&1 ($<$1\%)\\%
\hspace{3mm}50-54&5 (3\%)&27 (5\%)\\%
\hspace{3mm}55-59&11 (6\%)&54 (10\%)\\%
\hspace{3mm}60-64&43 (22\%)&123 (23\%)\\%
\hspace{3mm}65-69&72 (37\%)&219 (40\%)\\%
\hspace{3mm}$\geq$70&64 (33\%)&120 (22\%)\\%
\hspace{3mm}Missing&25&43\\%
\multicolumn{3}{l}{\hspace{-3mm}\textbf{PSA, ng/mL}}\\%
\hspace{3mm}$<$3&14 (8\%)&60 (11\%)\\%
\hspace{3mm}3-$<$5&26 (14\%)&135 (26\%)\\%
\hspace{3mm}5-$<$10&42 (23\%)&165 (31\%)\\%
\hspace{3mm}$\geq$10&103 (56\%)&166 (32\%)\\%
\hspace{3mm}Missing&36&61\\%
\midrule%
\multicolumn{3}{c}{\textbf{Whole Slide Images}}\\%
\midrule%
\textbf{\textit{n\tnote{*}}}&544&1,660\\%
\textbf{Physical slides}&311&856\\%
\multicolumn{3}{l}{\hspace{-2mm}\textbf{Gleason score}}\\%
\hspace{3mm}3 + 3&0 (0\%)&3 ($<$1\%)\\%
\hspace{3mm}3 + 4&17 (6\%)&189 (23\%)\\%
\hspace{3mm}3 + 5&3 (1\%)&15 (2\%)\\%
\hspace{3mm}4 + 3&86 (30\%)&231 (28\%)\\%
\hspace{3mm}4 + 4&112 (40\%)&216 (27\%)\\%
\hspace{3mm}4 + 5&53 (19\%)&103 (13\%)\\%
\hspace{3mm}5 + 3&0 (0\%)&1 ($<$1\%)\\%
\hspace{3mm}5 + 4&12 (4\%)&34 (4\%)\\%
\hspace{3mm}5 + 5&0 (0\%)&19 (2\%)\\%
\hspace{3mm}Missing&28&45\\%
\multicolumn{3}{l}{\hspace{-2mm}\textbf{ISUP}}\\%
\hspace{3mm}1&0 (0\%)&3 ($<$1\%)\\%
\hspace{3mm}2&17 (6\%)&189 (23\%)\\%
\hspace{3mm}3&86 (30\%)&231 (28\%)\\%
\hspace{3mm}4&115 (41\%)&232 (29\%)\\%
\hspace{3mm}5&65 (23\%)&156 (19\%)\\%
\hspace{3mm}Missing&28&45\\\bottomrule%
\end{tabular}}
\newcommand{\tableMetrics}{\begin{tabular}{@{}lccccc@{}}%
\toprule%
\textbf{Cohort}&\textbf{Type}&\textbf{AUC}&\textbf{Cohen's kappa}&\textbf{Sensitivity}&\textbf{Specificity}\\%
\midrule%
STHLM3&Internal&0.96 (0.94, 0.99)&0.8 (0.69, 0.9)&0.88 (0.78, 0.98)&0.95 (0.91, 0.98)\\%
SUH&Internal&0.98 (0.95, 1.0)&0.8 (0.63, 0.96)&1.0 (1.0, 1.0)&0.84 (0.7, 0.97)\\%
AMU&External&0.92 (0.86, 0.97)&0.42 (0.27, 0.6)&1.0 (1.0, 1.0)&0.49 (0.33, 0.63)\\%
MUL&External&0.89 (0.83, 0.94)&0.53 (0.39, 0.65)&0.89 (0.8, 0.97)&0.67 (0.56, 0.77)\\%
SCH&External&0.95 (0.87, 0.99)&0.71 (0.4, 0.93)&0.73 (0.43, 1.0)&0.96 (0.89, 1.0)\\%
\midrule%
Overall&Internal&0.97 (0.95, 0.99)&0.81 (0.72, 0.89)&0.92 (0.85, 0.98)&0.93 (0.89, 0.96)\\%
Overall&External&0.9 (0.86, 0.93)&0.55 (0.45, 0.64)&0.9 (0.84, 0.96)&0.7 (0.63, 0.76)\\\bottomrule%
\end{tabular}}
\newcommand{\tablePathologistPairwiseKappaAvg}{\begin{tabular}{@{}lc@{}}%
\toprule%
\textbf{Rater}&\textbf{Cohen's kappa}\\%
\midrule%
\textbf{Our model}&\textbf{0.66 (0.57, 0.74)}\\%
Pathologist 1&0.62 (0.52, 0.7)\\%
\textbf{Lead pathologist}&\textbf{0.61 (0.51, 0.7)}\\%
Pathologist 3&0.61 (0.5, 0.7)\\%
Pathologist 4&0.58 (0.46, 0.68)\\%
Pathologist 5&0.57 (0.45, 0.67)\\%
Pathologist 6&0.56 (0.45, 0.65)\\%
Pathologist 7&0.54 (0.43, 0.64)\\%
Pathologist 8&0.52 (0.4, 0.63)\\%
Pathologist 9&0.35 (0.22, 0.52)\\\bottomrule%
\end{tabular}}
\newcommand{\tableScannerConsistency}{\begin{tabular}{@{}lccccc@{}}%
\toprule%
\textbf{Scanner}&\textbf{Aperio}&\textbf{Grundium}&\textbf{Hamamatsu}&\textbf{Philips}&\textbf{Average}\\%
\midrule%
\textbf{Aperio}&-&0.97 (0.82, 1.00)&0.97 (0.83, 1.00)&0.93 (0.79, 1.00)&0.96 (0.87, 0.99)\\%
\textbf{Grundium}&0.97 (0.82, 1.00)&-&0.93 (0.77, 1.00)&0.97 (0.84, 1.00)&0.95 (0.87, 0.99)\\%
\textbf{Hamamatsu}&0.97 (0.83, 1.00)&0.93 (0.77, 1.00)&-&0.90 (0.74, 0.97)&0.93 (0.81, 0.99)\\%
\textbf{Philips}&0.93 (0.79, 1.00)&0.97 (0.84, 1.00)&0.90 (0.74, 0.97)&-&0.93 (0.80, 0.99)\\\bottomrule%
\end{tabular}}
\newcommand{\tableBorderlineAnalysis}{\begin{tabular}{@{}lcccccccc@{}}%
\toprule%
&&\multicolumn{3}{c}{\textbf{True Negatives}}&\multicolumn{3}{c}{\textbf{False Positives}}&\\%
\textbf{Cohort}&\textbf{Type}&n&Borderline&\%&n&Borderline&\%&\textit{p-value}\textsuperscript{*}\\%
\midrule%
STHLM3&Internal&452&22&5\%&24&10&42\%&\textbf{<0.001}\\%
SUH&Internal&26&1&4\%&5&2&40\%&0.06\\%
MUL&External&55&7&13\%&27&9&33\%&\textbf{0.038}\\%
SCH&External&43&7&16\%&2&2&100\%&\textbf{0.036}\\%
\midrule%
Overall&Internal&478&23&5\%&29&12&41\%&\textbf{<0.001}\\%
Overall&External&98&14&14\%&29&11&38\%&\textbf{0.008}\\\bottomrule%
\end{tabular}}

\title{Finding Holes: Pathologist Level Performance Using AI for Cribriform Morphology Detection in Prostate Cancer}
\renewcommand{\headeright}{}
\renewcommand{\undertitle}{}
\renewcommand{\shorttitle}{Using AI for Cribriform Morphology Detection in Prostate Cancer}
\date{October 13, 2025}

\author[1]{Kelvin Szolnoky}
\author[2,3]{Anders Blilie}
\author[1]{Nita Mulliqi}
\author[4]{Toyonori Tsuzuki}
\author[5]{Hemamali Samaratunga}
\author[1]{Matteo Titus}
\author[1]{Xiaoyi Ji}
\author[1,6]{Sol Erika Boman}
\author[2]{Einar Gudlaugsson}
\author[7,8]{Svein Reidar Kjosavik}
\author[9]{José Asenjo}
\author[10]{Marcello Gambacorta}
\author[10]{Paolo Libretti}
\author[11]{Marcin Braun}
\author[12]{Radzis\l{}aw Kordek}
\author[12]{Roman \L{}owicki}
\author[13,14]{Brett Delahunt}
\author[15]{Kenneth A. Iczkowski}
\author[16]{Theo van der Kwast}
\author[17]{Geert J. L. H. van Leenders}
\author[18]{Katia R. M. Leite}
\author[19]{Chin-Chen Pan}
\author[2,20,21]{Emiel Adrianus Maria Janssen}
\author[1]{Martin Eklund}
\author[14]{Lars Egevad}
\author[22]{Kimmo Kartasalo}

\affil[1]{Department of Medical Epidemiology and Biostatistics, Karolinska Institutet, Stockholm, Sweden}
\affil[2]{Department of Pathology, Stavanger University Hospital, Stavanger, Norway}
\affil[3]{Faculty of Health Sciences, University of Stavanger, Stavanger, Norway}
\affil[4]{Department of Surgical Pathology, School of Medicine, Aichi Medical University, Nagoya, Japan}
\affil[5]{Aquesta Uropathology and University of Queensland, Brisbane, Queensland, Australia}
\affil[6]{Department of Molecular Medicine and Surgery, Karolinska Institutet, Stockholm, Sweden}
\affil[7]{The General Practice and Care Coordination Research Group, Stavanger University Hospital, Stavanger, Norway}
\affil[8]{Department of Global Public Health and Primary Care, Faculty of Medicine, University of Bergen, Bergen, Norway}
\affil[9]{Department of Pathology, SYNLAB, Madrid, Spain}
\affil[10]{Department of Pathology, SYNLAB, Brescia, Italy}
\affil[11]{Department of Pathology, Chair of Oncology, Medical University of Lodz, Lodz, Poland}
\affil[12]{1st Department of Urology, Medical University of Lodz, Lodz, Poland}
\affil[13]{Malaghan Institute of Medical Research, Wellington, New Zealand}
\affil[14]{Department of Oncology and Pathology, Karolinska Institutet, Stockholm, Sweden}
\affil[15]{Department of Pathology and Laboratory Medicine, University of California - Davis Health, Sacramento, CA, USA}
\affil[16]{Laboratory Medicine Program and Princess Margaret Cancer Center, University Health Network, University of Toronto, Toronto, ON, Canada}
\affil[17]{Department of Pathology, Erasmus MC, University Medical Center, Rotterdam, the Netherlands}
\affil[18]{Department of Urology, Laboratory of Medical Research, University of São Paulo Medical School, Sao Paulo, Brazil}
\affil[19]{Department of Pathology and Laboratory Medicine, Taipei Veterans General Hospital, Taipei, Taiwan}
\affil[20]{Department of Chemistry, Bioscience and Environmental Engineering, University of Stavanger, Stavanger, Norway}
\affil[21]{Institute for Biomedicine and Glycomics, Griffith University, Brisbane, Queensland, Australia}
\affil[22]{Department of Medical Epidemiology and Biostatistics, SciLifeLab, Karolinska Institutet, Stockholm, Sweden}

\makeatletter
\newcommand{\theauthor}{\@author}
\makeatother

\hypersetup{
	pdftitle={\thetitle},
	pdfsubject={},
	pdfauthor={Kelvin Szolnoky},
	pdfkeywords={},
}

\begin{document}
\begin{refsection}

\maketitle
\clearpage

\begin{abstract}

	\textbf{Background:} Cribriform morphology in prostate cancer is a histological feature that indicates poor prognosis
	and contraindicates active surveillance. However, it remains underreported and subject to significant interobserver
	variability amongst pathologists. We aimed to develop and validate an AI-based system to improve cribriform pattern
	detection.

	\textbf{Methods:} We created a deep learning model using an EfficientNetV2-S encoder with multiple instance learning
	for end-to-end whole-slide classification. The model was trained on 640 digitised prostate core needle biopsies from
	430 patients, collected across three cohorts. It was validated internally (261 slides from 171 patients) and externally
	(266 slides, 104 patients from three independent cohorts). Internal validation cohorts included laboratories or
	scanners from the development set, while external cohorts used completely independent instruments and laboratories.
	Annotations were provided by three expert uropathologists with known high concordance. Additionally, we conducted an
	inter-rater analysis and compared the model's performance against nine expert uropathologists on 88 slides from the
	internal validation cohort.

	\textbf{Results:} The model showed strong internal validation performance (AUC: 0.97, 95\% CI: 0.95-0.99; Cohen's
	kappa: 0.81, 95\% CI: 0.72-0.89) and robust external validation (AUC: 0.90, 95\% CI: 0.86-0.93; Cohen's kappa: 0.55,
	95\% CI: 0.45-0.64). In our inter-rater analysis, the model achieved the highest average agreement (Cohen's kappa:
	0.66, 95\% CI: 0.57-0.74), outperforming all nine pathologists whose Cohen's kappas ranged from 0.35 to 0.62.

	\textbf{Conclusion:} Our AI model demonstrates pathologist-level performance for cribriform morphology detection in
	prostate cancer. This approach could enhance diagnostic reliability, standardise reporting, and improve treatment
	decisions for prostate cancer patients.

\end{abstract}

\section{Introduction}
Cribriform morphology in prostate cancer indicates increased metastatic potential, and is associated with adverse
outcomes and increased mortality~\cite{kweldam_cribriform_2015,russo_oncological_2023}. The term cribriform comes from
the Latin \textit{cribrum}, meaning \textit{sieve}, which describes its appearance where malignant epithelial cells
form sheets punctured by sieve-like spaces~\cite{epstein_2014_2016, van_der_kwast_isup_2021}. By definition, cribriform
morphology is classified as at least Gleason pattern 4~\cite{epstein_2014_2016}. In core needle biopsies, the
prevalence of cribriform morphology ranges from 4\% (for Gleason 3+4) up to 21\% (for higher grade
tumors)~\cite{osiecki_presence_2024,flammia_cribriform_2020, egevad_interobserver_2023}. Given its prognostic value,
the presence of cribriform morphology now contraindicates active surveillance strategies in prostate cancer
management~\cite{ericson_diagnostic_2020}.

Despite this clinical importance, cribriform morphology remains underreported in routine
practice~\cite{osiecki_presence_2024}. This creates gaps in patient risk stratification. Furthermore, like Gleason
grading, identifying cribriform patterns shows substantial interobserver variability and requires specialist expertise
for consistent identification~\cite{egevad_interobserver_2023}. These diagnostic challenges are compounded by
increasing workload pressures in pathology departments. Rising case volumes and declining number of specialists are
stretching resources~\cite{markl_number_2021}.

While AI solutions have emerged to address workload challenges, current approaches fail to fully meet the spectrum of
diagnostic needs. Many AI models for prostate cancer focus solely on Gleason score~\cite{rabilloud_deep_2023}. However,
comprehensive pathological reporting requires additional features beyond Gleason scoring. An effective AI solution must
recognise and report multiple pathological features from a biopsy, with cribriform morphology detection being
particularly important.

Currently, no study has sufficiently validated an AI-based system for cribriform detection. This study therefore aims
to develop and validate an AI model for the automatic detection of cribriform morphology in prostate core needle
biopsies.

\section{Materials and Methods}

\begin{figure}
	\centering
	\includegraphics[width=\linewidth]{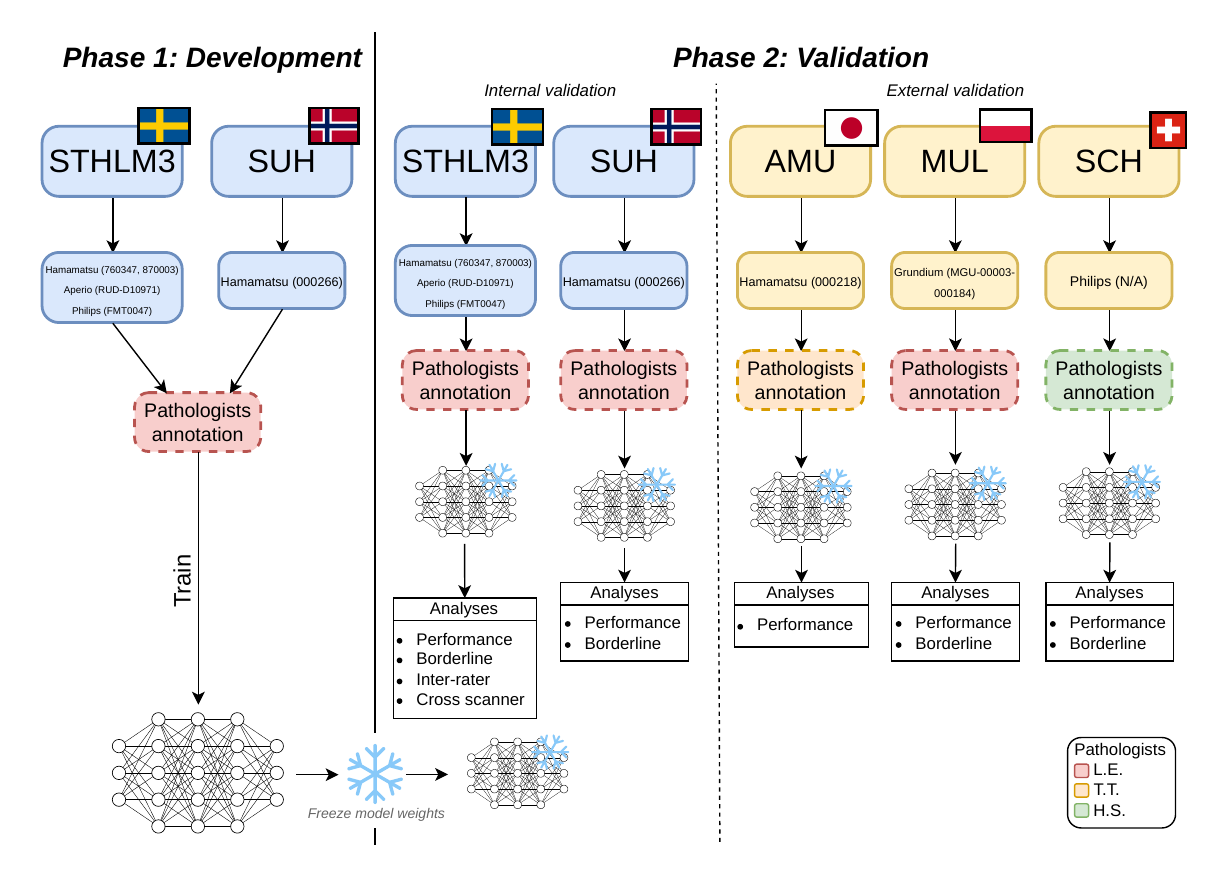}

	\caption{Overview of the study design. Phase 1 (Development) used subsets of the STHLM3 and SUH cohorts for model training.
		Phase 2 (Validation) included internal validation on reserved STHLM3/SUH data and external validation on three independent
		cohorts (AMU, MUL, SCH). Slides were digitised on scanners from multiple vendors and annotated by three pathologists. Numbers
		in parentheses indicate scanner serial numbers. Serial numbers for the scanners used at SCH are unavailable, but these scanners
		are distinct from those used in the other cohorts. No scanners in the external cohorts were present in the training data.
		Performance evaluation included standard metrics (AUC, Cohen's kappa, sensitivity, specificity), inter-rater analysis comparing our model with nine pathologists, cross-scanner reproducibility assessment, and borderline case analysis.\\
		\textit{Definition of abbreviations:} AUC = Area under the receiver operating characteristic curve.}
	\label{fig:study-flowchart}
\end{figure}

We conducted a retrospective study in two phases: (1) model development and (2) validation
(Figure~\cref{fig:study-flowchart}). The study protocol has been published~\cite{mulliqi_development_2025}.

\subsection{Data and Participants}
We digitised formalin-fixed paraffin-embedded (FFPE), haematoxylin and eosin-stained prostate core needle biopsy slides
from six cohorts: the Stockholm3 (STHLM3) trial~\cite{gronberg_prostate_2015}, Capio S:t Göran Hospital, Sweden (STG),
Stavanger University Hospital, Norway (SUH), Aichi Medical University, Japan (AMU), Medical University of Lodz, Poland
(MUL), and Synlab Switzerland (SCH). Complete information about the cohorts -- including collection dates,
participants, and sampling methods -- can be found in the protocol~\cite{mulliqi_development_2025}. The slides were
digitised using eight scanner instruments from four different vendors, including Philips (STHLM3, SCH), Grundium (MUL),
Hamamatsu (AMU, STHLM3, SUH), and Aperio (STHLM3, STG). Some slides were scanned multiple times using different
scanners. Please refer to Table 2 in the protocol for details regarding slide digitisation across
cohorts~\cite{mulliqi_development_2025}.

Parts of the STHLM3, STG, and SUH cohorts were used for training and tuning the model during phase 1 (model
development), while a portion of data from these cohorts was reserved for internal validation during phase 2 (model
validation). The AMU, MUL, and SCH cohorts were used entirely for external validation during phase 2. The validation
datasets used during phase 2 were completely independent from the development process in phase 1 and only used for
validation once. In other words, after phase 1, the model remained entirely fixed (frozen) throughout phase 2 without
any adjustments. We defined validation cohorts as ``internal'' when their laboratory and/or specific scanner instrument
had been included in the development set, while ``external'' cohorts contained samples from physical scanner
instruments and laboratories that were completely independent from those used in development. All data partitions used
to separate development from validation sets were grouped at the patient level to prevent data leakage, ensuring that
slides from the same patient never appeared in both training and validation datasets.

\subsection{Outcome}

Cribriform morphology was defined per ISUP 2021 consensus as confluent malignant epithelial cells with multiple
glandular lumina visible at low power (x10 objective), without intervening stroma or mucin between glandular
structures~\cite{van_der_kwast_isup_2021}. Cribriform growth was annotated irrespective of whether it was invasive
(within acinar adenocarcinoma) or non-invasive (intraductal carcinoma). This was justified by both forms often being
assessed and reported together for prognostication and treatment planning, a practice supported by the 2019 ISUP
consensus~\cite{van_leenders_2019_2020}.

To minimise interobserver variability, we established a reference standard based on annotations from the lead
pathologist (L.E.) or other experienced uropathologists (H.S., T.T.) whose concordance has been quantified in earlier
studies~\cite{egevad_interobserver_2023}. To reduce the annotation burden for the reference standard pathologists,
non-reference standard pathologists initially reviewed cases with Gleason pattern 4 to identify suspect slides with
cribriform morphology. These preliminary annotations were used to upsample suspect cribriform cases for subsequent
reference standard annotation. A non-reference standard pathologist was defined as one whose concordance to the lead
pathologist (L.E.) is unknown. Non-reference standard annotations were not used to assess model performance.

For the STHLM3 and STG cohorts, a collection of 700 slides containing Gleason pattern 4 was assessed and annotated by
the lead pathologist. In the other cohorts (SUH, MUL, and SCH), initial annotations were made by non-reference standard
pathologists. A sample, with positive cases upsampled, was re-labelled by a reference standard pathologist. Detailed
annotation protocols for each cohort are provided in Table~\cref{tab:sampling} and the
protocol~\cite{mulliqi_development_2025}.

For STHLM3 and STG, pixel-level annotations for glands representing cribriform morphology were made. For the other
cohorts, only slide-level labels were annotated. When establishing the reference standard, L.E. (on STHLM3, SUH, and
MUL) and H.S. (on SCH) also indicated cases they considered borderline cribriform. The term borderline was used for
cases with features suggestive of cribriform growth that did not fully meet established morphological criteria. This
category was intended to capture diagnostically difficult cases to permit statistical analyses on this specific
substratum.

For a subset of the STHLM3 internal validation data, we also have annotations from the nine expert uropathologists
included in an earlier interobserver reproducibility study~\cite{egevad_interobserver_2023}.

\subsection{Model Development}
We extracted smaller images, referred to as patches, from each whole slide image (WSI) for input into the model. Each
patch measured 256 by 256 pixels at 1 $\mu$m per pixel (10x magnification) and overlapped with neighbouring patches by
50\% both vertically and horizontally. Patches with tissue covering less than 10\% of the image were discarded based on
tissue segmentation masks. An in-house segmentation model built on UNet++ with a ResNeXt-101 (32x4d) encoder was used
to create the tissue segmentation masks~\cite{boman_impact_2025}.

We developed a multiple instance learning (MIL) model using an EfficientNetV2-S neural network backbone to detect
cribriform morphology. The model processes WSIs by using the extracted patches, treating each slide as a bag of
patches. These patches are processed through the neural network to extract patch-level features, which are aggregated
via a gated attention mechanism to create slide-level features. To enhance generalisability, we implemented extensive
data augmentation techniques. The final model utilised an ensemble of 10 models trained during 10-fold
cross-validation, and used test-time augmentation for final predictions. Further details are available in the
supplement (Section~\cref{sec:supp-methods}).

\subsection{Statistical Analysis}
All analyses were prespecified~\cite{mulliqi_development_2025}. We performed analyses at both individual cohort levels
and aggregated internal and external cohort levels. Model performance was evaluated using receiver operating
characteristic (ROC) curves and the area under the ROC curve (AUC). An operating point of 0.5 was used for binary
classification. We calculated sensitivity and specificity. We also measured the agreement between the model and the
reference standard using Cohen's kappa. For glass slides that were digitised multiple times in the STHLM3 cohort,
performance metrics were calculated using only the original WSI that was annotated by the pathologist, rather than
including all digital copies of the same physical slide. The 95\% confidence intervals (CIs) for all metrics were
calculated from nonparametric bootstrapping using 1,000 bootstrap samples. We created visual calibration plots to
assess model calibration.

Using annotations on a subset of STHLM3 data from nine pathologists and our model, we conducted an inter-rater
variability analysis to compare the model's performance against expert pathologists. For each rater, including our
model, we calculated the mean pairwise Cohen's kappa coefficient against the other pathologists to quantify agreement
levels. Furthermore, we conducted sensitivity analyses to evaluate cross-scanner reproducibility by calculating the
pairwise Cohen's kappa between model predictions on different digital scans of the same glass slides. This analysis
used slides from the STHLM3 cohort that had been digitised multiple times using scanners from four vendors (Aperio,
Grundium, Hamamatsu, Philips). Lastly, in an exploratory analysis, we quantified the prevalence of ``borderline'' cases
in both true negative and false positive groups using annotations from L.E. and H.S. In analyses not specifically
focused on borderline cases, these were classified as negative.

\section{Results}

\subsection{Dataset characteristics}

\begin{table}
	\centering
	\makebox[\textwidth][c]{%
		\begin{threeparttable}
			\caption{Patient and slide characteristics across all cohorts, showing demographic and clinical data.}%
			\label{tab:one}
			\tableOne

			\begin{tablenotes}
				\footnotesize
				\item[*] Total number of whole slide images (digital copies of physical slides). This may exceed the number of physical slides when slides from a cohort were scanned multiple times on different scanners.
				\item \textit{Definition of abbreviations:} PSA = Prostate specific antigen; ISUP = International Society of Urological
				Pathology Grade.
			\end{tablenotes}
		\end{threeparttable}
	}
\end{table}

Patient and slide characteristics are summarised in Table~\cref{tab:one,tab:one-cribriform}. The study included a total
of 705 patients: 430 in the training set, 171 in the internal validation set, and 104 in the external validation set
(Table~\cref{tab:one-split}). Training was done on 1,280 WSIs from 640 physical slides. The internal validation cohorts
included 211 physical slides from STHLM3 and 50 from SUH. The external validation cohorts contained 137 slides from MUL
and 56 from SCH. The prevalence of cribriform pattern was higher in the external validation set (35\%, $n=94$) compared
to training (24\%, $n=155$) and internal validation sets (24\%, $n=62$). The most common age interval was 65-69 years,
comprising 40\% of patients. Gleason score and ISUP grade distributions were relatively consistent across training,
internal validation, and external validation sets.

\subsection{Model Performance}

\begin{table}
	\centering

	\begin{threeparttable}
		\caption{Performance metrics across all cohorts, including AUC, Cohen's kappa, sensitivity, and specificity values with 95\% confidence intervals. Type indicates the cohort's validation status.}
		\tableMetrics
		\label{tab:metrics}

		\begin{tablenotes}
			\footnotesize
			\item \textit{Definition of abbreviations:} AUC = Area under the receiver operating characteristic curve.
		\end{tablenotes}
	\end{threeparttable}

\end{table}

\begin{figure}
	\centering
	\begin{subfigure}{0.47\textwidth}
		\centering
		\includegraphics[width=\linewidth]{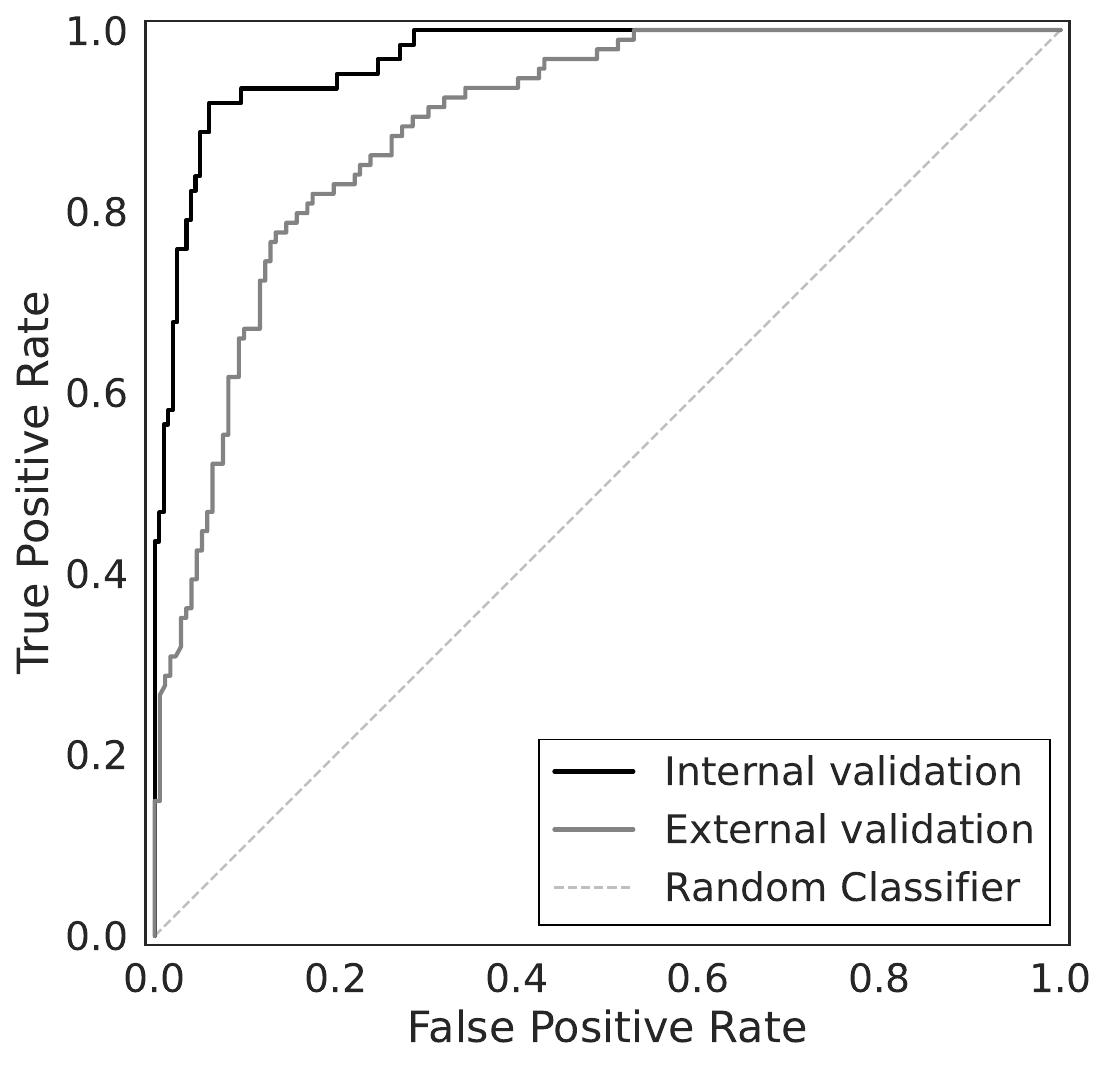}
		\caption{}
		\label{fig:roc-agg}
	\end{subfigure}\\
	\begin{subfigure}{0.47\textwidth}
		\centering
		\includegraphics[width=\linewidth]{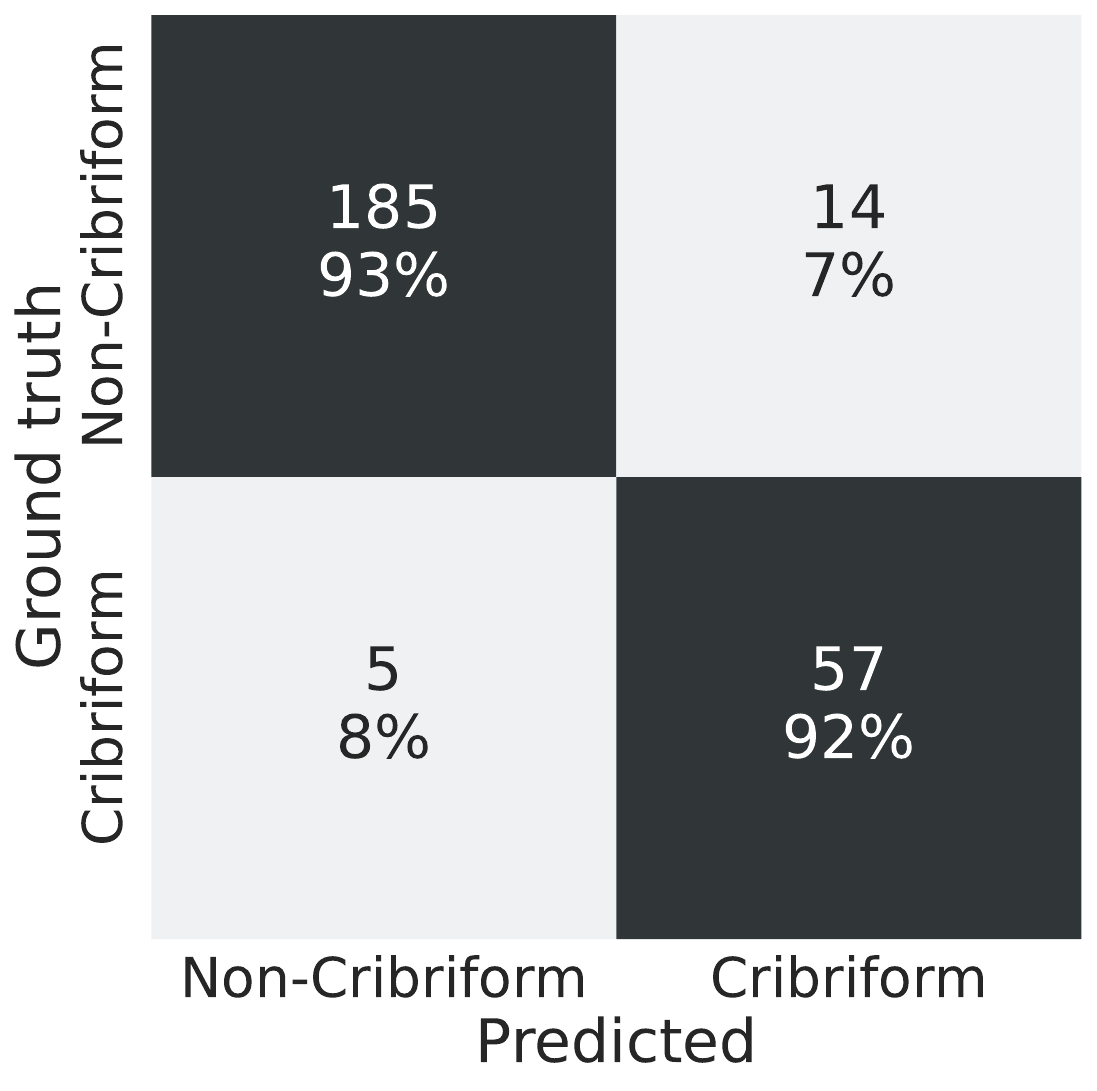}
		\caption{}
		\label{fig:cm-internal}
	\end{subfigure}%
	\begin{subfigure}{0.47\textwidth}
		\centering
		\includegraphics[width=\linewidth]{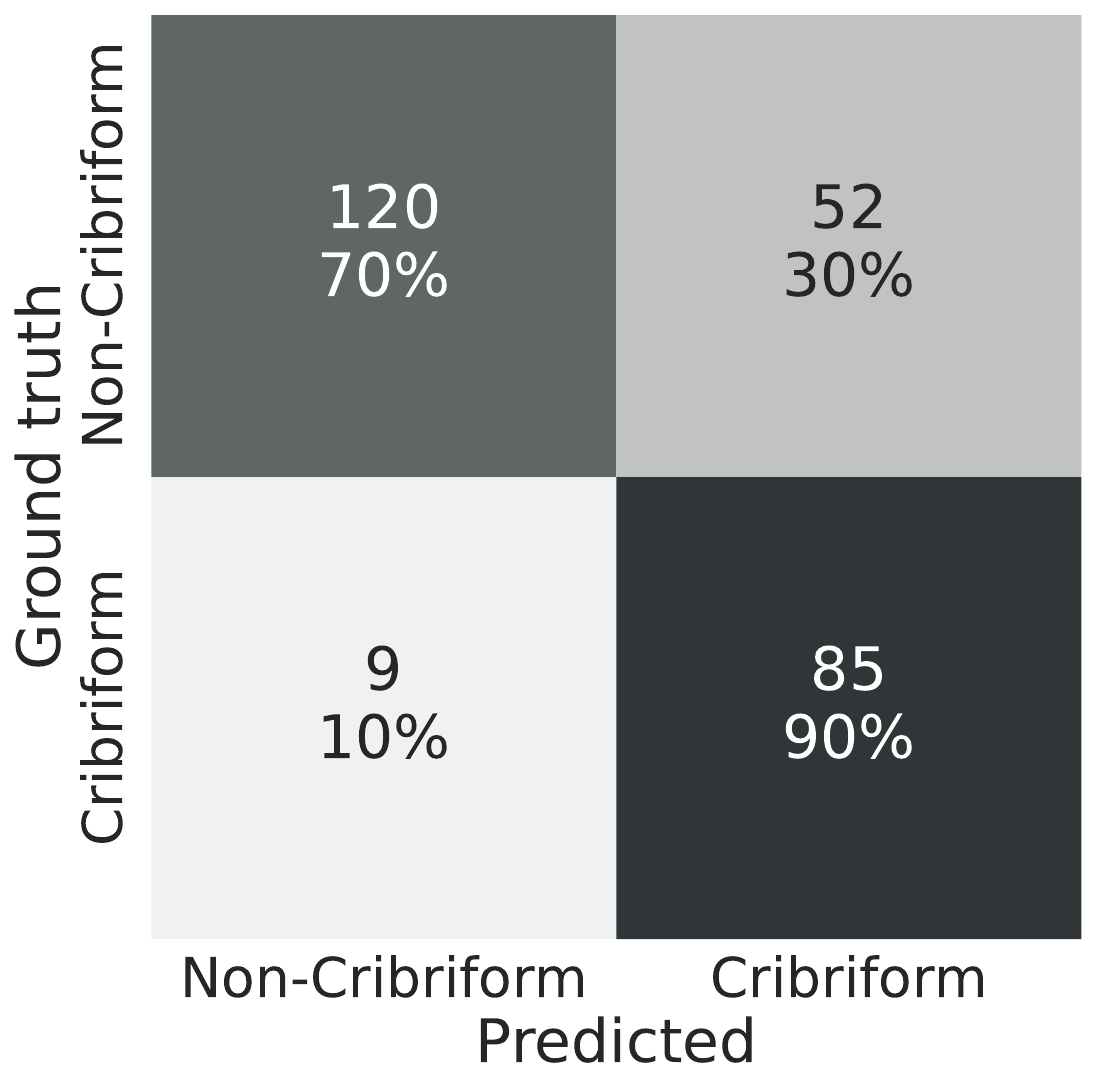}
		\caption{}
		\label{fig:cm-external}
	\end{subfigure}

	\caption{
		(a) Receiver operating characteristic curves showing model performance on internal and external validation sets.
		(b) Confusion matrix on predictions for the internal validation set (STHLM3 and SUH).
		(c) Confusion matrix on predictions for the external validation set (AMU, MUL, and SCH).
	}
	\label{fig:performance-main}
\end{figure}

On the internal validation set (STHLM3 and SUH cohorts), our deep learning model demonstrated an AUC of 0.97 (95\% CI:
0.95, 0.99) and a Cohen's kappa of 0.81 (95\% CI: 0.72, 0.89), with a sensitivity of 0.92 (95\% CI: 0.85, 0.98) and
specificity of 0.93 (95\% CI: 0.89, 0.96). For external validation (AMU, MUL, and SCH), the model achieved an AUC of
0.90 (95\% CI: 0.86, 0.93) and a Cohen's kappa of 0.55 (95\% CI: 0.45, 0.64), with a sensitivity of 0.90 (95\% CI:
0.84, 0.96) and specificity of 0.70 (95\% CI: 0.63, 0.76). The ROC curves and confusion matrices illustrating these
performance differences are presented in Figure~\cref{fig:roc-agg,fig:roc-cohort}, while AUC, Cohen's kappa,
sensitivity, and specificity for all included cohorts are presented in Table~\cref{tab:metrics}.

Examining individual cohorts (Table~\cref{tab:metrics}), performance varied across datasets. STHLM3 achieved an AUC of
0.96 (95\% CI: 0.94, 0.99) and a Cohen's kappa of 0.80 (95\% CI: 0.69, 0.90), while SUH demonstrated similar results
with an AUC of 0.98 (95\% CI: 0.95, 1.0) and a similar Cohen's kappa of 0.80 (95\% CI: 0.63, 0.96). Performance in
external validation cohorts was more variable. The SCH cohort maintained results comparable to internal validation,
with an AUC of 0.95 (95\% CI: 0.87, 0.99) and a Cohen's kappa of 0.71 (95\% CI: 0.40, 0.93). However, while the AMU and
MUL cohorts preserved good discriminative ability with AUCs of 0.92 (95\% CI: 0.86, 0.97) and 0.89 (95\% CI: 0.83,
0.94) respectively, their agreement metrics were notably lower, with Cohen's kappa values of 0.42 (95\% CI: 0.27, 0.60)
for AMU and 0.53 (95\% CI: 0.39, 0.65) for MUL.

The model showed good calibration internally, with predicted probabilities closely matching observed cribriform
morphology (Figure~\cref{fig:calibration-agg,fig:calibration-cohort}). In external validation, calibration deviated --
especially at intermediate probabilities -- leading to overdiagnosis and reduced specificity at the same operating
point (0.5) used on the internal validation sets (Figure~\cref{fig:cm-external,fig:cm-cohort}).

In our exploratory analysis of borderline cases (Table \cref{tab:borderline}), using annotations from L.E. (on STHLM3,
SUH, and MUL) and H.S. (SCH), we found a significantly higher proportion of borderline cases among false positive cases
(38\%) compared to true negative cases (14\%; Fisher's exact test, $p=0.008$). In the cross-scanner reproducibility
analysis all scanners showed high concordance, with pairwise agreement ranging from 0.90 to 0.97. Scanner specific
results for the cross-scanner reproducibility analysis are presented in the supplementary material
(Section~\cref{sec:supp-results}).

\subsection{Comparison with Pathologists}

\begin{figure}
	\centering
	\includegraphics[width=0.8\linewidth]{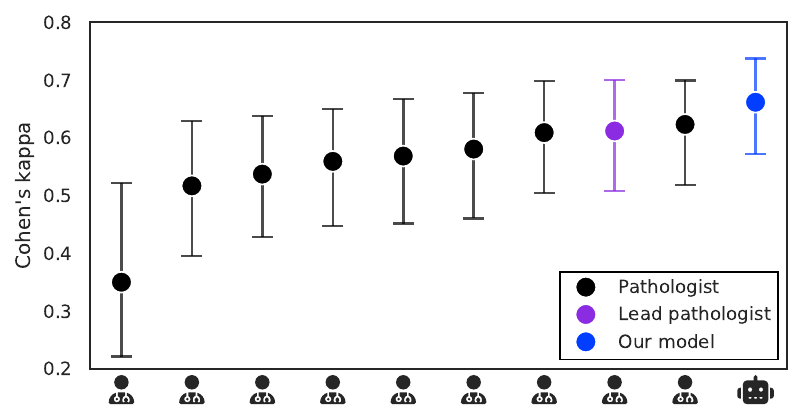}
	\caption{Pathologist concordance analysis comparing the agreement between our model (robot icon) and nine pathologists (physician icon),
		showing mean pairwise Cohen's kappa values. For each rater, including our model, the mean pairwise Cohen's kappa was calculated
		against the other pathologists only (the model was excluded from this average calculation). The whiskers indicate the 95\% confidence interval.
		For exact values see Table~\cref{tab:pathologist-concordance}.}
	\label{fig:path-concordance}
\end{figure}

In the inter-rater analysis (Figure~\cref{fig:path-concordance} and Table~\cref{tab:pathologist-concordance}) we
compared the model with nine pathologists on a subset of 88 slides from the STHLM3 validation cohort. Using the lead
pathologist's (L.E.) annotations as a reference, 43 slides were positive for the cribriform pattern. The model achieved
the highest average pairwise Cohen's kappa of 0.66 (95\% CI: 0.57, 0.74). This exceeded the performance of all nine
pathologists, whose average pairwise Cohen's kappa values ranged from 0.35 (95\% CI: 0.22, 0.52) to 0.62 (95\% CI:
0.52, 0.70). The lead pathologist, who annotated the training data, ranked 3rd with an average pairwise Cohen's kappa
of 0.61 (95\% CI: 0.51, 0.7).

\section{Discussion}
In this study, we developed and validated a deep-learning model to detect cribriform morphology in prostate cancer
biopsies. Our model demonstrated strong discriminative performance across both internal and external cohorts, achieving
high agreement with experienced pathologists' annotations and, notably, the highest average agreement scores when
compared against nine pathologists. However, in some external cohorts, a shift in calibration was observed, leading to
an overdiagnosis of cribriform patterns. Even so, the model maintained acceptable performance levels comparable to
those of a pathologist. Roughly 40\% of the model's false positive cases were considered to be borderline cribriform
cases by the annotating pathologists. These findings suggest the model's potential value as a screening tool for
identifying high-risk regions within slides and helping to prioritise the most diagnostically challenging cases for
expert pathologist review.

A strength of our study is the comprehensive validation strategy, which employed both internal and external validation
cohorts alongside inter-rater and cross-scanner reproducibility analyses. However, some limitations must be
acknowledged. Specificity fell from 93\% internally to 70\% externally, indicating calibration and cohort shift.
Inter-observer variability likely contributes, as prior work shows only moderate agreement among expert uropathologists
assessing cribriform morphology (mean Cohen's kappa $\approx$ 0.56)~\cite{egevad_interobserver_2023}. Additionally,
both the cross-scanner reproducibility and pathologist concordance analyses were conducted on internal validation sets,
potentially overestimating performance. These analyses were enabled by the availability of multi-rater annotations and
repeatedly scanned slides from prior studies exclusively on the STHLM3 cohort.

To our knowledge, this is the first study to comprehensively validate an AI model specifically for cribriform pattern
detection in prostate cancer. Previous research has primarily focused on Gleason grading or tumour
detection~\cite{rabilloud_deep_2023}. Two earlier studies featured models for cribriform detection, but showed only
modest results and lacked external multi-cohort validation~\cite{ambrosini_automated_2020,silva-rodriguez_going_2020}.
Our approach advances this work by developing a model that has been validated across multiple external, international
cohorts and by comparing model performance directly against multiple pathologists.

The accurate detection of cribriform morphology represents one of the crucial decision points in treatment planning for
prostate cancer patients. This prognostically significant pattern, though often overlooked in practice, directly
influences risk stratification and treatment selection. Our model attempts to address this challenge by providing
support for pathologists and enhancing diagnostic consistency. For pathologists confronting mounting caseloads, the
model could offer assistance by highlighting regions of interest, streamlining workflows, and supporting diagnostic
decisions. Improved reliability in cribriform detection could translate to better-informed treatment assessments for
patients. Future research should focus on improving external calibration and conducting prospective clinical
validation.

\section{Conclusion}
Our deep learning model demonstrates robust performance for automated cribriform morphology detection in prostate
cancer, with performance comparable to experienced pathologists. This approach could enhance diagnostic reliability,
standardise reporting of this prognostically important feature, and potentially improve treatment decisions for
prostate cancer patients.

\printbibliography
\end{refsection}

\section*{Ethical considerations}

The study is conducted in agreement with the Declaration of Helsinki. The data were retrieved in one or more rounds at
each of the participating international sites between 1 May 2012 and 1 May 2024. All data were deidentified at each
site and provided to Karolinska Institutet in anonymised format. The centralised collection of patient samples from the
international sites to Karolinska Institutet was approved by the Swedish Ethical Review Authority (permit 2019-05220).
The following local approvals were provided to cover the data collection at each site: AMU (permit 2023-074 for the AMU
cohort), Stockholm regional ethics committee (permits 2012/572-31/1, 2012/438-31/3, and 2018/845-32 for the STG and
STHLM3 cohorts), the Bioethics Committee at the Medical University of Lodz (permit RNN/295/19/KE for the MUL cohort),
and the Regional Committee for Medical and Health Research Ethics (REC) in Western Norway (permits REC/Vest 80924, REK
2017/71 for the SUH cohort). For the SCH cohort, ethical approval was waived by the respective local institutional
review boards due to the retrospective usage of fully deidentified prostate specimens, and the data collection under
the waiver was approved by the Swedish Ethical Review Authority (permit 2019-05220). Written informed consent was
provided by the participants in the STHLM3 dataset.

\section*{Acknowledgements}
The computations were made possible through the National Academic Infrastructure for Supercomputing in Sweden (NAISS) and the
Swedish National Infrastructure for Computing (SNIC) at C3SE partially funded by the Swedish Research Council through
grant agreement no. 2022-06725 and no. 2018-05973, and by the supercomputing resource Berzelius provided by the National
Supercomputer Centre at Linköping University and the Knut and Alice Wallenberg Foundation.

We want to thank Carin Cavalli-Björkman for assistance with scanning and database support. We would also like to thank
Silja Kavlie Fykse and Desmond Mfua Abono for scanning in Stavanger. We would like to acknowledge the patients who
participated in the STHLM3 diagnostic study and the OncoWatch and NordCaP projects and contributed the clinical
information that made this study possible.

A.B. received a grant from the Health Faculty at the University of Stavanger, Norway. M.E. received funding from the
Swedish Research Council, Swedish Cancer Society, Swedish Prostate Cancer Society, Nordic Cancer Union, Karolinska
Institutet, and Region Stockholm. K.K. received funding from the SciLifeLab \& Wallenberg Data Driven Life Science
Program (KAW 2024.0159), David and Astrid Hägelen Foundation, Instrumentarium Science Foundation, KAUTE Foundation,
Karolinska Institute Research Foundation, Orion Research Foundation and Oskar Huttunen Foundation. L.E. received
funding from the Swedish Cancer Foundation (23 2641 Pj) and the Stockholm Cancer Society (234053).

\section*{Author's Contributions}

K.S., N.M., X.J., S.E.B. and K.K. developed the AI models. A.B., E.G., S.R.K., J.A., M.G., P.L., M.B., R.K., R.\L{}.,
B.D., H.S., T.T., E.A.M.J., K.A.I., T.v.d.K, G.J.L.H.v.L, K.R.M.L., C.P., and L.E. collected, assessed and curated
clinical datasets. N.M., X.J., K.S., S.E.B. and K.K. contributed to digitization, pre-processing and management of
whole slide image data. K.S. conducted the statistical analyses. K.S. and K.K. analyzed and interpreted the study
results. N.M. and K.K. acquired, optimized and maintained computing platforms. A.B., M.E. and K.K. acquired funding.
L.E., M.E. and K.K. conceived of the study. K.K. takes responsibility for the integrity and accuracy of the analysis in
this study. K.S., drafted the manuscript. All authors reviewed, edited and approved the manuscript.

\newpage
\renewcommand{\shorttitle}{Supplement: Using AI for Cribriform Morphology Detection in Prostate Cancer}

\begin{refsection}
\appendix
\thispagestyle{empty}

\begin{center}
	\hrule height 2pt\vspace*{0.15in}%
	{\LARGE\bfseries SUPPLEMENTARY MATERIAL}\\[1em]
	{\Large\bfseries\sc\thetitle\par}\vspace{0.15in}%
	\hrule height 2pt\vspace*{2em}%
	{\theauthor}\\[3em]
	{\thedate}
\end{center}

\newpage

\setcounter{table}{0}
\setcounter{figure}{0}
\renewcommand{\thetable}{A\arabic{table}}
\renewcommand{\thefigure}{A\arabic{figure}}

\vspace*{\fill}

\section{\centering\Large\bfseries Figures and Tables}

\vspace*{\fill}
\newpage

\begin{table}
	\centering
	\caption{Labelling and sampling methodology across different cohorts, sampling strategies, and the annotating pathologist (reference standard).}
	\label{tab:sampling}
	\makebox[\textwidth][c]{%
		\begin{threeparttable}
			\begin{tabular}{@{}lccccc@{}}
				\toprule
				\textbf{Cohort} & \textbf{Initial sampling\tnote{*}} & \textbf{Initial annotator} & \textbf{Second sampling\tnote{*}}                                                  & \textbf{Reference standard} \\ \midrule
				STHLM3/STG      & 701 slides containing GP 4         & L.E.                       & N/A                                                                                & L.E.                        \\
				SUH             & 332 slides containing GP 4         & A.B.                       & 120\textsuperscript{+}/40\textsuperscript{borderline}/40\textsuperscript{-} slides & L.E.                        \\
				AMU             & 73 slides containing GP 4          & T.T.                       & N/A                                                                                & T.T.                        \\
				MUL             & 276 slides containing GP 4         & A.B.                       & 74\textsuperscript{+}/63\textsuperscript{-} slides                                 & L.E.                        \\
				SCH             & 56  slides containing GP 4         & Site pathologists          & 12\textsuperscript{+}/6\textsuperscript{-} blocks                                  & H.S.                        \\ \bottomrule
			\end{tabular}
			\begin{tablenotes}
				\small
				\item \textit{Definition of abbreviations:} GP 4 = Gleason pattern 4.
				\item[*] To enhance statistical power and reduce the annotation burden for the reference standards, a two-stage enrichment sampling strategy was employed.
				Initially, a non-reference standard pathologist annotated slides containing Gleason pattern 4. Subsequently, slides were resampled based on these preliminary
				annotations to enrich for potential cribriform patterns before final annotation by the reference standard pathologist. In the \textit{second sampling} column,
				superscript symbols indicate the initial pathologist's assessment for cribriform and how sampling was done based off of these annotations. In the STHLM3, STG, and AMU
				cohorts the initial annotator was the reference standard, i.e. no second round of annotations was needed.
			\end{tablenotes}
		\end{threeparttable}
	}
\end{table}

\begin{table}
	\centering
	\caption{Hyperparameters for the patch level model.}
	\begin{tabular}{@{}lp{10cm}@{}}
		\toprule
		\textbf{Hyperparameter} & \textbf{Value}                                                                                                                                                                                                  \\ \midrule
		Encoder                 & EfficientNetV2-S                                                                                                                                                                                                \\
		Initial weights         & Weights from Gleason scoring encoder                                                                                                                                                                            \\
		Loss function           & Binary cross entropy loss (weighted)                                                                                                                                                                            \\
		Optimiser               & AdamW                                                                                                                                                                                                           \\
		Learning rate           & OneCycleLR scheduler (starting at $1 \cdot 10^{-5}$, peaking at $1 \cdot 10^{-4}$ after 1 epoch, and finally decreasing to $1 \cdot 10^{-6}$ following a cosine annealing schedule)                             \\
		Weight decay            & $1\cdot10^{-2}$                                                                                                                                                                                                 \\
		Batch size              & 64                                                                                                                                                                                                              \\
		Precision               & bfloat16                                                                                                                                                                                                        \\
		Train augmentations     & Random: crop, horizontal and vertical flip, 90 degrees rotation, colour jitter, gamma, tone curve, grey scale, unsharp mask or guassian blur, ISO noise, gaussian noise, multiplicative noise, JPEG compression \\
		\bottomrule
	\end{tabular}
	\label{tab:tile-hyperparameters}
\end{table}

\begin{table}
	\centering
	\caption{Hyperparameters for the slide level model.}
	\begin{tabular}{@{}lp{10cm}@{}}
		\toprule
		\textbf{Hyperparameter} & \textbf{Value}                                                                                                                                                                                                  \\ \midrule
		Encoder                 & EfficientNetV2-S                                                                                                                                                                                                \\
		Initial weights         & Cribriform patch level weights                                                                                                                                                                                  \\
		Loss function           & Binary cross entropy loss (weighted)                                                                                                                                                                            \\
		Optimiser               & RAdam                                                                                                                                                                                                           \\
		Learning rate           & $3\cdot10^{-5}$ (constant)                                                                                                                                                                                      \\
		Weight decay            & $1\cdot10^{-5}$                                                                                                                                                                                                 \\
		Batch size              & 1                                                                                                                                                                                                               \\
		Max bag size            & 2200                                                                                                                                                                                                            \\
		Precision               & bfloat16                                                                                                                                                                                                        \\
		Train augmentations     & Random: crop, horizontal and vertical flip, 90 degrees rotation, colour jitter, gamma, tone curve, grey scale, unsharp mask or gaussian blur, ISO noise, gaussian noise, multiplicative noise, JPEG compression \\
		Test augmentations      & Random: horizontal and vertical flip, 90 degrees rotation                                                                                                                                                       \\
		\bottomrule
	\end{tabular}
	\label{tab:slide-hyperparameters}
\end{table}

\begin{table}
	\centering
	\begin{threeparttable}
		\caption{Patient and slide characteristics stratified by dataset split (training, internal validation, and external validation).}
		\label{tab:one-split}
		\tableOneSplit

		\begin{tablenotes}
			\footnotesize
			\item[*] Total number of whole slide images (digital copies of physical slides). This may exceed the number of physical slides when slides from a cohort were scanned multiple times on different scanners.
		\end{tablenotes}
	\end{threeparttable}
\end{table}

\begin{table}
	\centering
	\begin{threeparttable}
		\caption{Patient and slide characteristics stratified by cribriform morphology status.}
		\label{tab:one-cribriform}
		\tableOneCribriform

		\begin{tablenotes}
			\footnotesize
			\item[*] Total number of whole slide images (digital copies of physical slides). This may exceed the number of physical slides when slides from a cohort were scanned multiple times on different scanners.
		\end{tablenotes}
	\end{threeparttable}
\end{table}

\begin{table}
	\caption{Mean pairwise Cohen's kappa values for our model and nine pathologists, evaluating 88 slides (43 annotated cribriform-positive by the lead pathologist) from the STHLM3 cohort. For each rater, including our model, the average was calculated against the pathologists only (the model was excluded from this average calculation). Values in parentheses indicate the 95\% confidence interval.}
	\centering
	\tablePathologistPairwiseKappaAvg
	\label{tab:pathologist-concordance}
\end{table}

\begin{table}
	\caption{Cross-scanner reproducibility analysis showing pairwise Cohen's kappa values between different scanner types for 71 slides (19 annotated cribriform-positive by the lead pathologist) from the STHLM3 validation set that were scanned on 4 different scanners. Values in parentheses indicate the 95\% confidence interval.}
	\centering
	\makebox[\textwidth][c]{%
		\tableScannerConsistency
	}
	\label{tab:scanner-cons}
\end{table}

\begin{table}
	\caption{Analysis of borderline cases comparing the prevalence of borderline cribriform morphology, as annotated by two experienced uropathologists, between true negative and false positive predictions. Type indicates the cohort's validation status.}
	\centering
	\makebox[\textwidth][c]{%
		\tableBorderlineAnalysis
	}
	\label{tab:borderline}

	\hspace{2.3em}\raggedright\footnotesize\textsuperscript{*} Fisher's exact test
\end{table}

\begin{figure}
	\centering
	\begin{subfigure}{0.26666666666666666\textwidth}
		\centering
		\includegraphics[width=\textwidth]{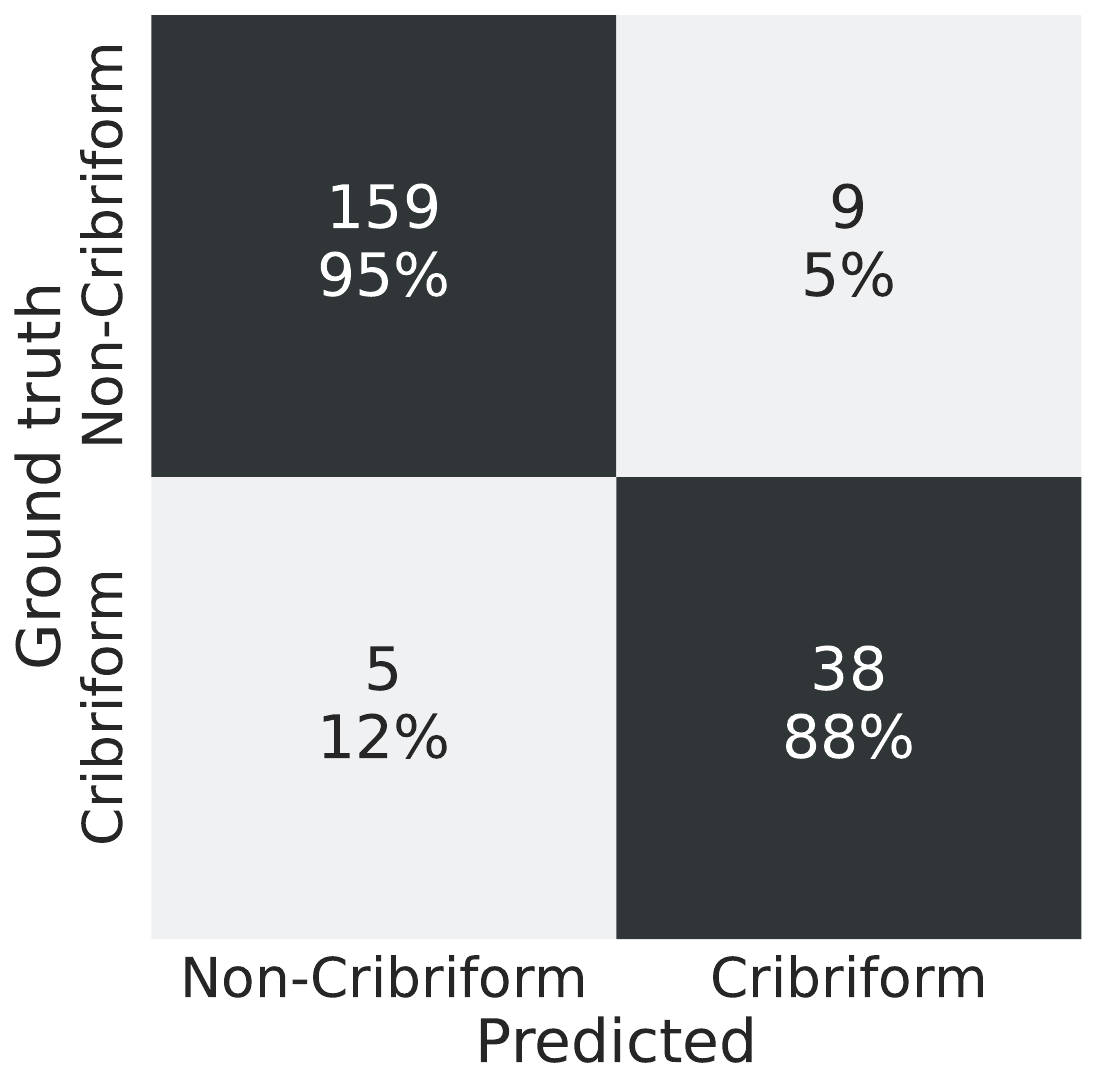}%
		\caption{}
	\end{subfigure}
	\hspace*{2em}
	\begin{subfigure}{0.26666666666666666\textwidth}
		\centering
		\includegraphics[width=\textwidth]{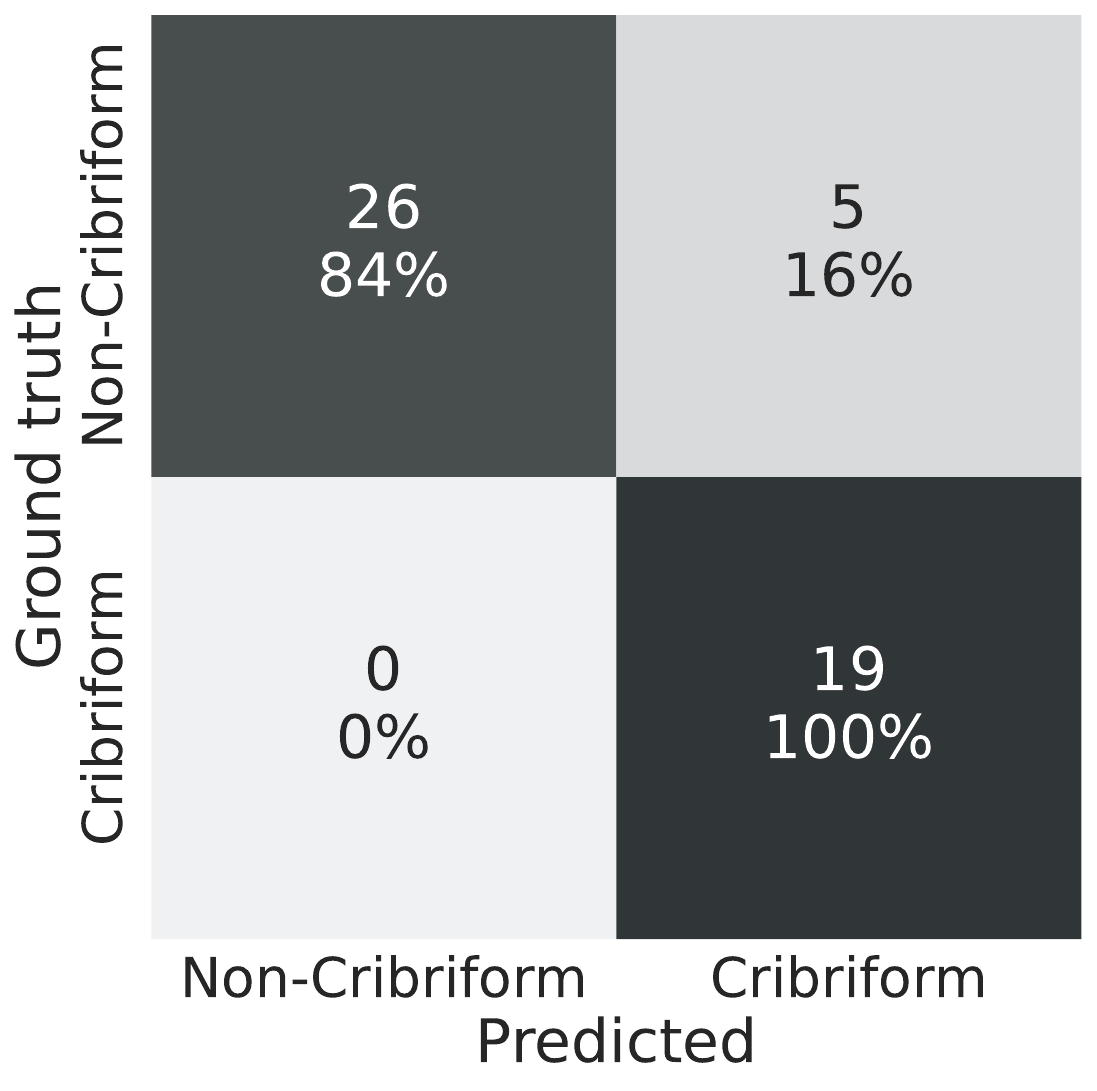}%
		\caption{}
	\end{subfigure}
	\hspace*{2em}
	\begin{subfigure}{0.26666666666666666\textwidth}
		\centering
		\includegraphics[width=\textwidth]{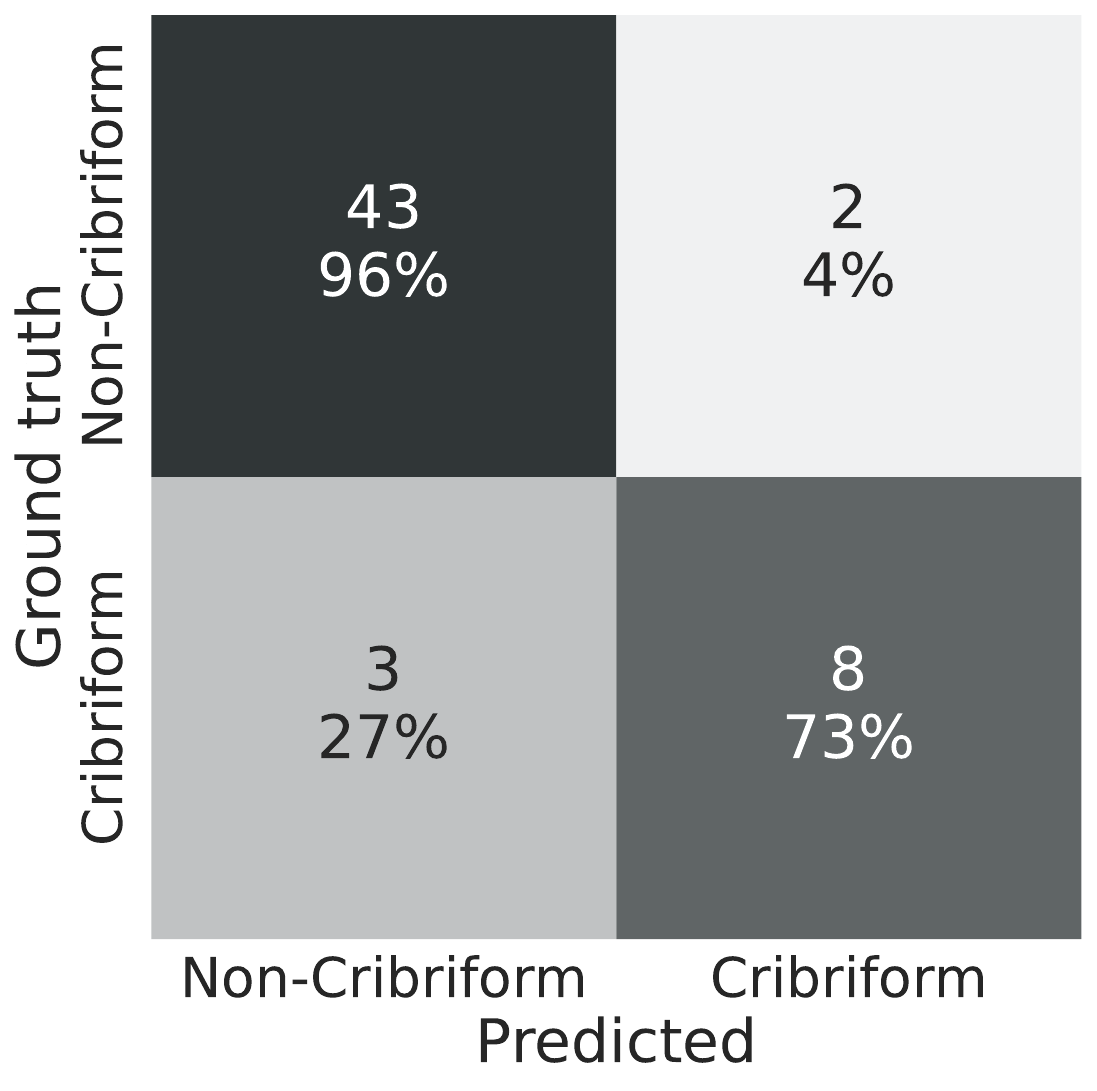}%
		\caption{}
	\end{subfigure}
	\\
	\vspace*{2em}
	\begin{subfigure}{0.26666666666666666\textwidth}
		\centering
		\includegraphics[width=\textwidth]{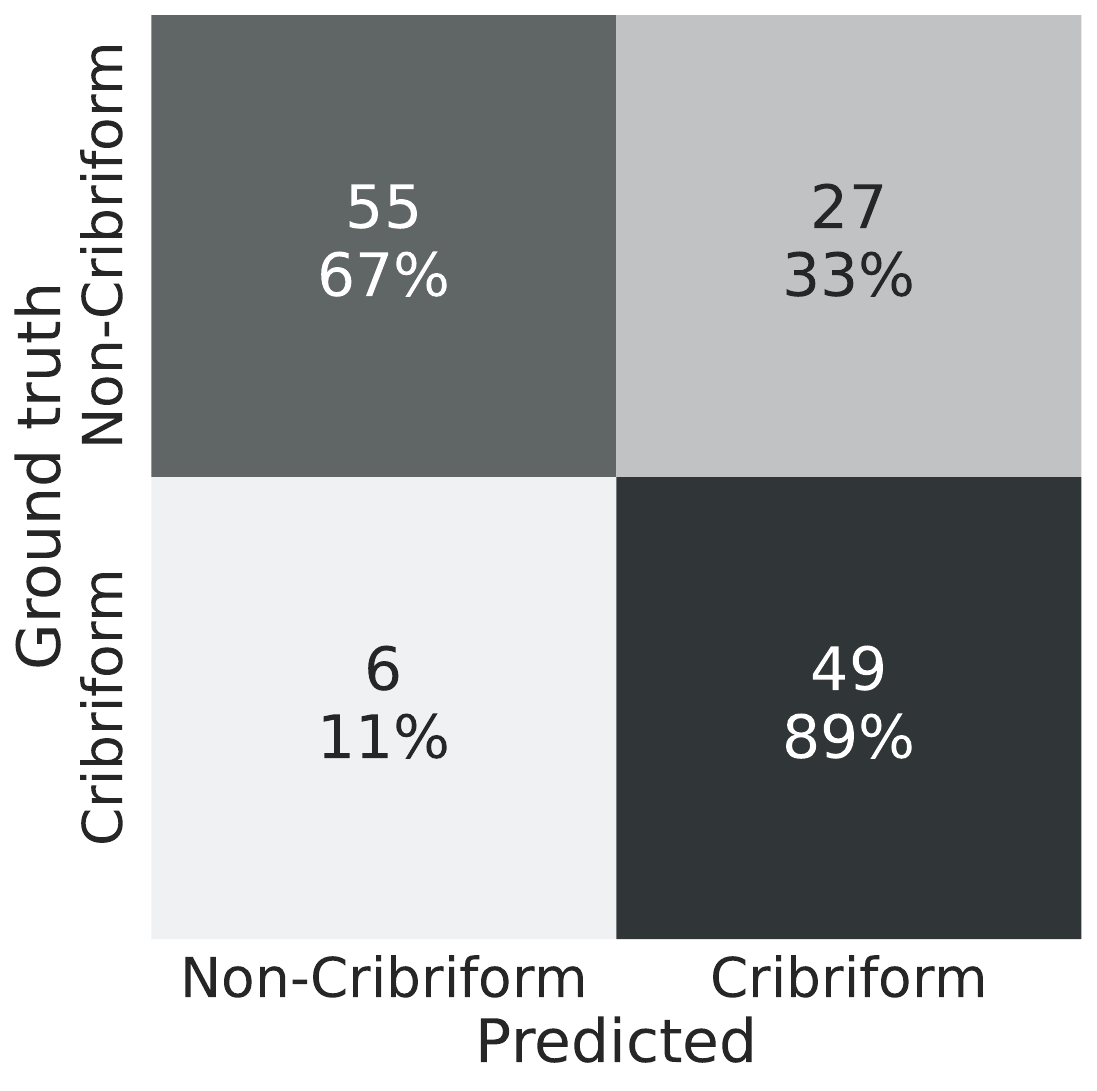}%
		\caption{}
	\end{subfigure}
	\hspace*{2em}
	\begin{subfigure}{0.26666666666666666\textwidth}
		\centering
		\includegraphics[width=\textwidth]{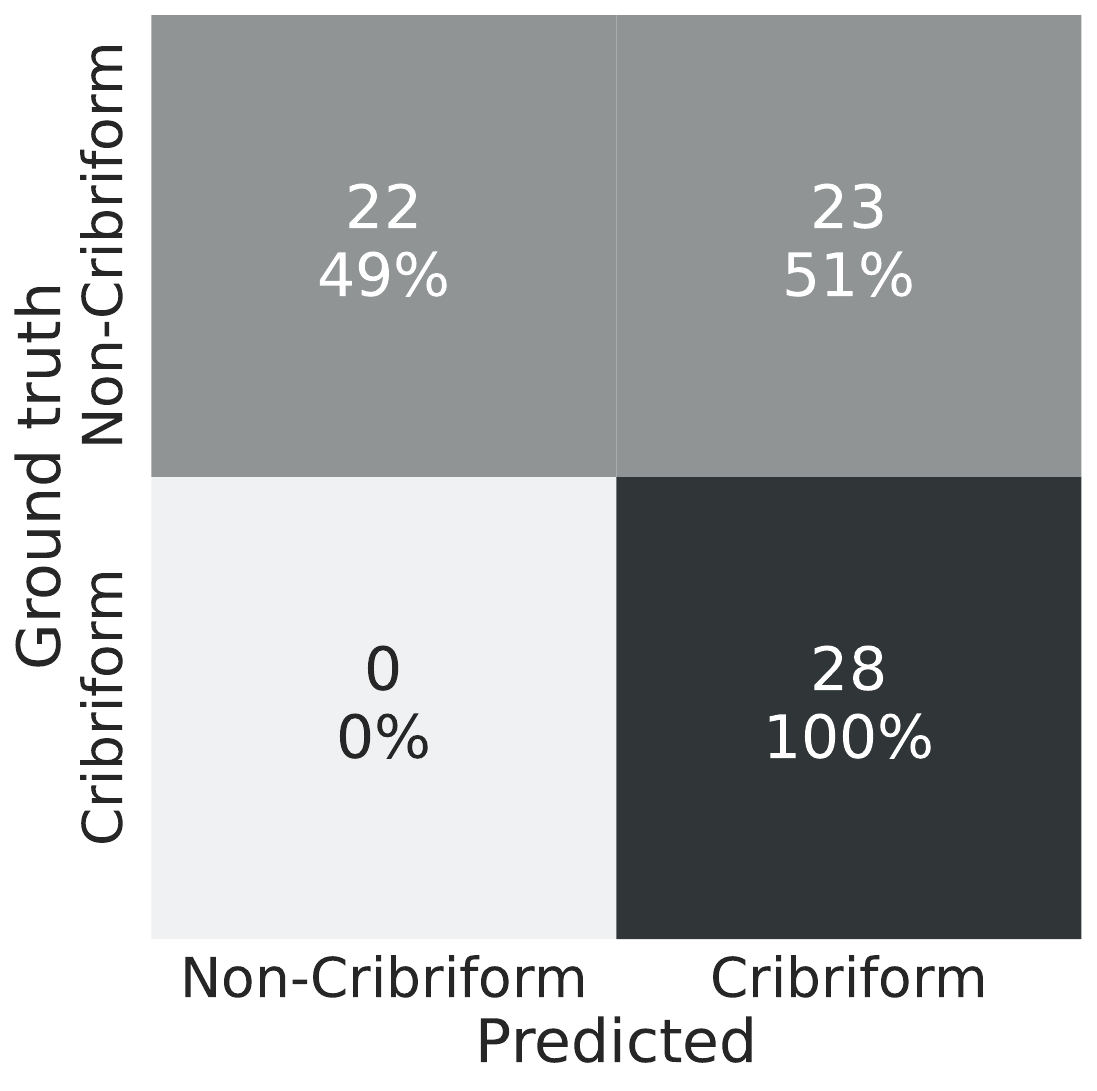}%
		\caption{}
	\end{subfigure}
	\caption{Confusion matrices on predictions for cohorts (a) STHLM3 (b) SUH (c) SCH (d) MUL (e) AMU}
	\label{fig:cm-cohort}
\end{figure}

\begin{figure}
	\centering
	\includegraphics[width=0.4\textwidth]{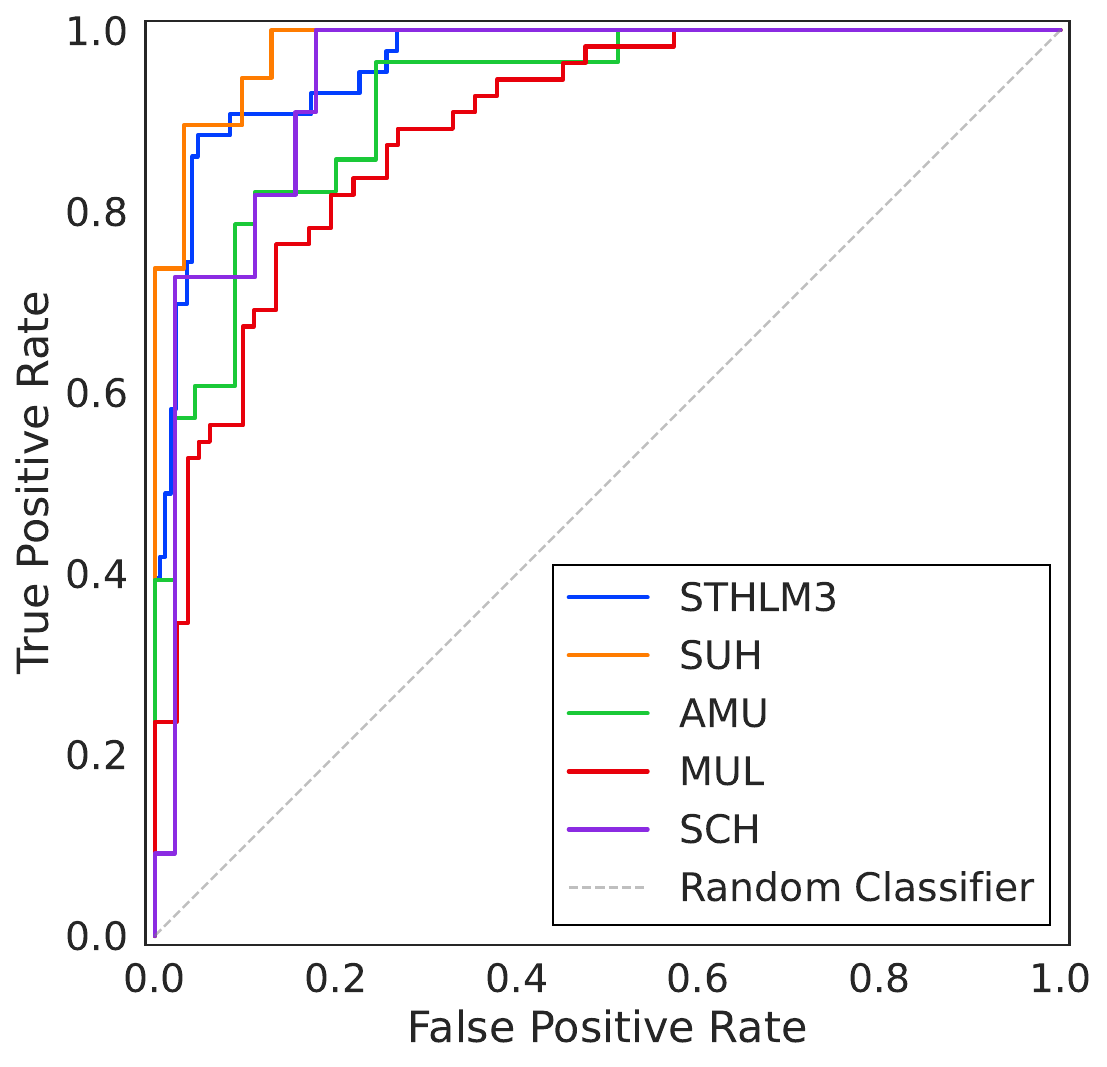}
	\caption{Receiver operating characteristic curves showing model performance for the different cohorts.}
	\label{fig:roc-cohort}
\end{figure}

\begin{figure}
	\centering
	\begin{subfigure}{0.4\textwidth}
		\centering
		\includegraphics[width=\textwidth]{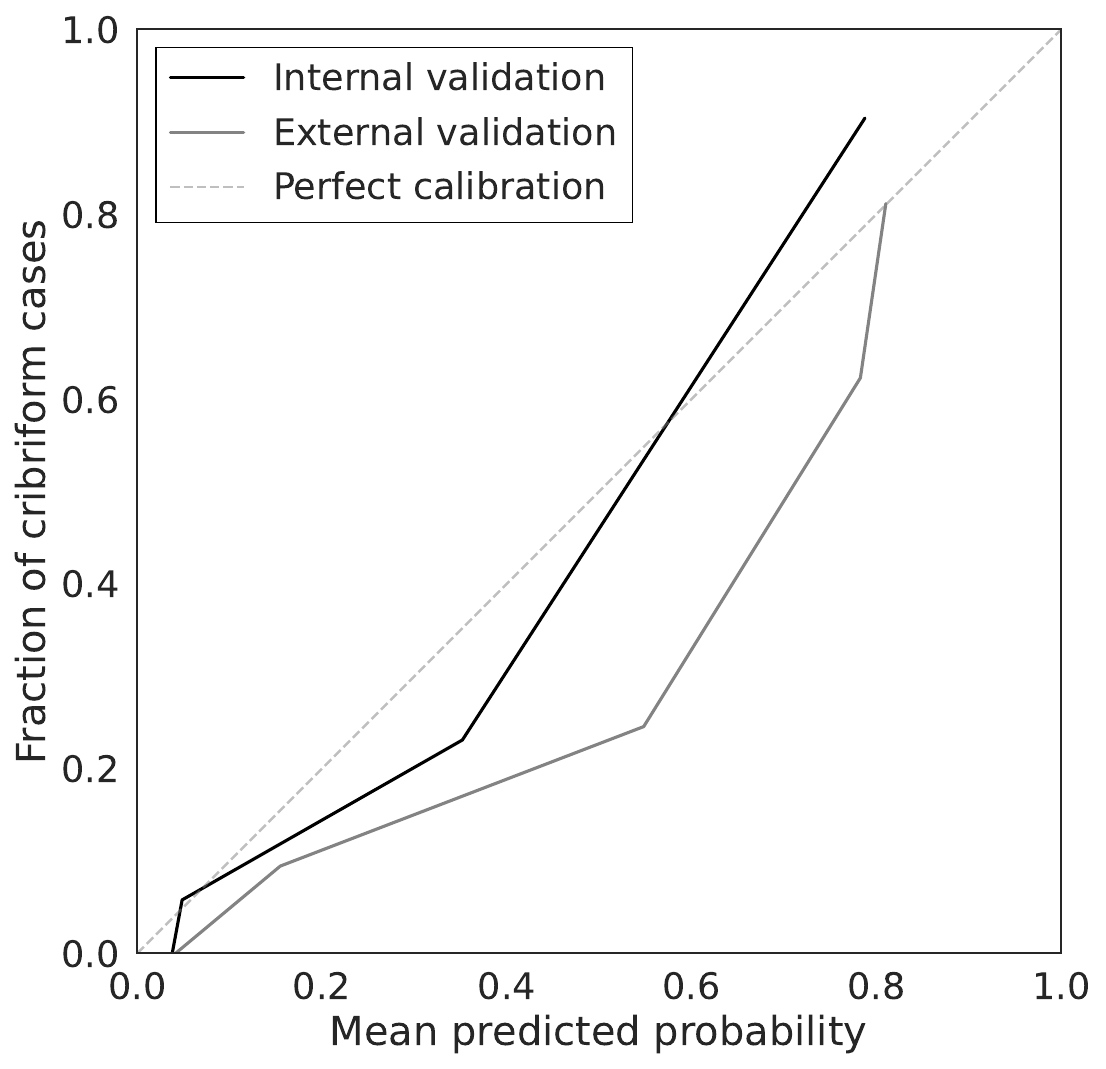}
		\caption{}
		\label{fig:calibration-agg}
	\end{subfigure}
	\hspace*{2em}
	\begin{subfigure}{0.4\textwidth}
		\centering
		\includegraphics[width=\textwidth]{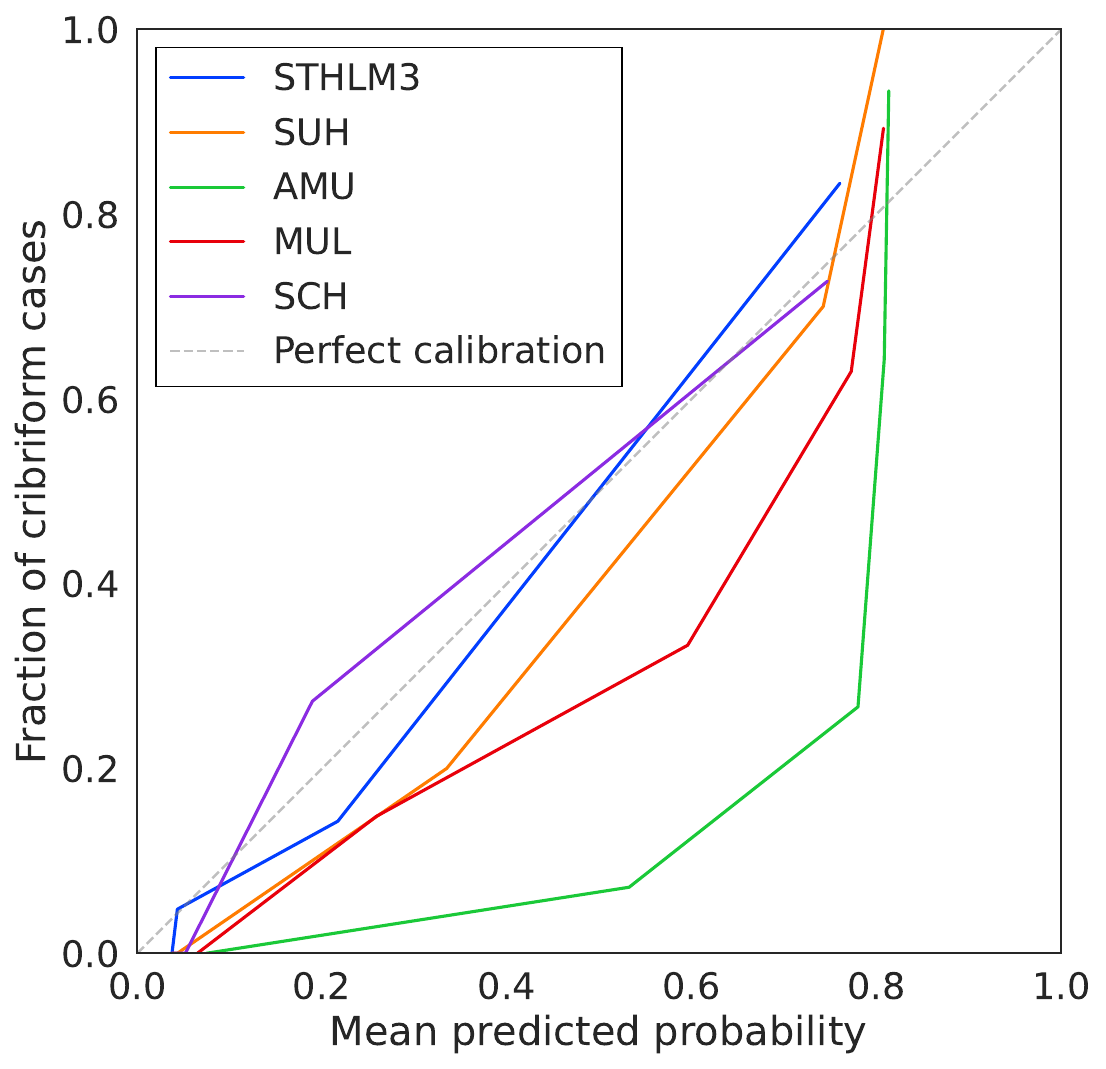}
		\caption{}
		\label{fig:calibration-cohort}
	\end{subfigure}

	\caption{(a) Calibration curves demonstrating the relationship between predicted probabilities and observed frequencies of cribriform morphology. (b) Calibration curves demonstrating the relationship between predicted probabilities and observed frequencies of cribriform morphology for the different cohorts.}
	\label{fig:calibration}
\end{figure}

\clearpage

\section{Materials and Methods}
\label{sec:supp-methods}

\subsection{Data Preparation}
For STHLM3 and STG, pixel-wise annotations were made. The lead pathologist created pixel-wise annotations (marking
cribriform regions) on only one digital version of each slide. Due to differences in how each scanner positioned the
slide during digitisation, these annotations could not be directly transferred to other digital versions of the same
slide. To address this limitation and increase our training data, we designed a simple phase correlation-based image
registration algorithm. For slides with multiple scans, we created binary tissue segmentation masks and used a Fast
Fourier Transform-based cross-correlation algorithm to align annotations across different digital versions of the same
physical glass slide.

\subsection{Model Development}
\begin{figure}[h!]
	\centering
	\includegraphics[width=\linewidth]{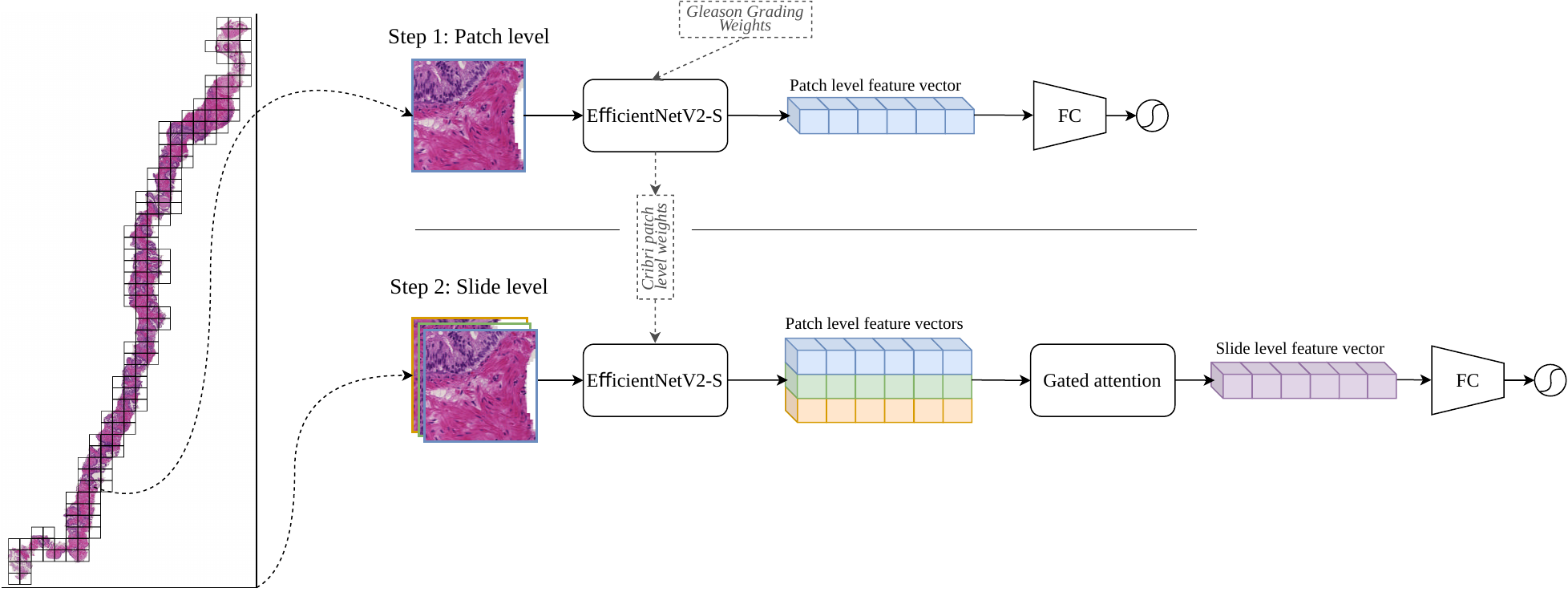}
	\caption{Model architecture illustrating the patch level and slide level classifiers for cribriform morphology detection in prostate cancer.}%
	\label{fig:model}%
	\raggedright\footnotesize{\textit{Definition of abbreviations:} FC = Fully connected layer.}
\end{figure}

We implemented a two-step transfer learning procedure to enhance performance, convergence speed, and generalisation.
The model architecture and transfer learning process are illustrated in Figure~\cref{fig:model}. This approach
furthermore enabled the incorporation of SUH cohort data during Step 2, as this dataset lacked pixel-level annotations
required for fully supervised training.

\textbf{Step 1: Fully Supervised Patch Level Classifier}. We created a patch level classifier using a convolutional
neural network. EfficientNetV2-S was chosen as the backbone due to its high performance and relatively low resource
usage~\cite{tan_efficientnetv2_2021}. To detect cribriform morphology at the patch level, we fine-tuned an
EfficientNetV2-S encoder, which was previously part of a multiple instance learning (MIL) model trained on a large
Gleason grading dataset~\cite{mulliqi_foundation_2025}. The feature vector was passed through a fully connected layer
with a sigmoid activation function, producing a probability score for cribriform morphology.

\textbf{Step 2: Weakly Supervised Slide Level Classifier}. We then developed a slide level classifier by transferring
the cribriform patch level encoder weights into a MIL architecture. The encoder weights were not frozen, allowing for
further fine-tuning during slide-level training. Bags of patches from a slide were processed through the encoder to
create patch level feature vectors. These vectors were pooled together using gated attention to form a single slide
level feature vector, which was then passed through a sequence of fully connected layers with normalisation,
activation, and dropout, followed by a sigmoid activation function to produce a slide level probability score.

\subsection{Training}
We used a binary cross-entropy loss function, weighted by the frequency of positive labels in the training data, with a
static probability threshold of 0.5 for classification. The AdamW optimiser was employed with a one cycle learning rate
scheduler for the patch level model (starting at $1 \cdot 10^{-5}$, peaking at $1 \cdot 10^{-4}$ after 1 epoch, and
finally decreasing to $1 \cdot 10^{-6}$) and a constant learning rate of $3\cdot10^{-5}$ for the slide level model. The
patch level classifier uses a weight decay of $1\cdot10^{-2}$ while the slide level model uses $1\cdot10^{-5}$. Data
augmentations included random cropping, vertical and horizontal flips, colour and brightness jitter, sharpening,
blurring, noise, JPEG compression, and random greyscale conversion. Furthermore, during training, for slides scanned
multiple times on different scanners, we randomised which digital scan of a biopsy slide to use on each epoch.
Hyperparameters are summarised in Tables~\cref{tab:tile-hyperparameters,tab:slide-hyperparameters}. The models were
trained for 8 and 32 epochs for the patch level and slide level respectively, with checkpoints every epoch, retaining
only the checkpoint with the highest non-weighted Cohen's kappa on the hold-out fold. We used 10-fold cross-validation
to evaluate the model during development. To avoid data leakage when transferring the Gleason grading weights, we
employed protocol-defined splits. Platt scaling was applied by fitting a logistic regression model to the held-out
folds in the training cross-validation.

For the patch level classifier, patches were labelled as cribriform-positive if they contained more than 2\% of
positively annotated tissue. Due to the constraints of needing pixel-wise annotations for the patch level classifier,
the classifier could only be trained on data from the STHLM3 and STG cohorts. For the slide level classifier, bags of
patches were labelled based on slide level annotations, enabling the classifier to be trained on the SUH cohort as
well. To further utilise pixel-wise annotations in the STHLM3 and STG data, bags of patches from STHLM3 and STG were
labelled as positive only if they contained a patch with cribriform morphology. For further regularisation, with the
50\% overlap that the patches were extracted with, we could construct two sets of non-overlapping patches per WSI. For
each forward pass of the model, one of these two non-overlapping sets was used. During validation, we used all
extracted patches from a slide.

\subsection{Inference}
Our final model utilised a 10-fold ensemble approach derived from the cross-validation folds during model development.
For inference, we extracted patches of size 256 by 256 pixels at a resolution of 1 $\mu$m per pixel, with 50\% overlap
between adjacent patches both vertically and horizontally. Patches were excluded if tissue content comprised less than
10\% of the image. All extracted patches were passed through the model. To enhance prediction robustness, we applied
test time augmentation with 5 iterations per ensemble model, using non-destructive transformations at a patch level,
such as flipping and rotation. The final prediction was generated through soft voting, averaging predictions across all
test time augmentation iterations and ensemble models.

\subsection{Software and Hardware}
Models and statistical analyses were implemented in Python 3.10 using Pytorch version 2.4 and Pytorch Lightning version
2.3. Support packages included Albumentations (1.4.12), LMDB (1.5.1), Numpy (2.1.3), Polars (1.14.0), Scipy (1.14.1),
Scikit-learn (1.5.2), Timm (1.0.11). Plots were made using Matplotlib (3.9.2) and Seaborn (0.13.2). Models were trained
using a single NVIDIA A100 80gb Tensor Core GPU. Inference was run on a single NVIDIA A100 40gb Tensor Core GPU. The
extracted patches were encoded as JPEGs and saved to the Lightning Memory-Mapped Database (LMDB) format, as this
allowed for efficient random reads which were needed in both phases of training. The resulting patch level and slide
level models had 20.2 million and 21.9 million trainable parameters, respectively. The final training took 10 hours per
data fold for a total of 100 GPU hours.

\section{Results}
\label{sec:supp-results}
\subsection{Cross-scanner Reproducibility}
When examining cross-scanner reproducibility (Table \cref{tab:scanner-cons}), we utilised 71 slides from the STHLM3
internal validation set that had all been scanned on scanners from 4 different vendors. The subset included 19 slides
that contained the cribriform pattern. The average pairwise Cohen's kappa values across scanners demonstrated high
consistency, with Aperio achieving the highest average agreement at 0.96 (95\% CI: 0.87, 0.99), followed by Grundium at
0.95 (95\% CI: 0.87, 0.99), while Hamamatsu and Philips showed average Cohen's kappa values of 0.93 (95\% CI: 0.78,
0.99) and 0.93 (95\% CI: 0.81, 0.99) respectively. The highest level of agreement was observed between Aperio and
Grundium scanners (Cohen's kappa: 0.97, 95\% CI: 0.80, 1.00) and between Aperio and Hamamatsu scanners (Cohen's kappa:
0.97, 95\% CI: 0.82, 1.00). The lowest agreement was found between Hamamatsu and Philips scanners with a Cohen's kappa
value of 0.90 (95\% CI: 0.73, 0.97).

\printbibliography

@article{ambrosini_automated_2020,
	title = {Automated Detection of Cribriform Growth Patterns in Prostate
	         Histology Images},
	author = {Ambrosini, Pierre and Hollemans, Eva and Kweldam, Charlotte F. and
	          Leenders, Geert J. L. H. Van and Stallinga, Sjoerd and Vos, Frans},
	date = {2020-09-10},
	journaltitle = {Scientific Reports},
	shortjournal = {Sci Rep},
	volume = {10},
	number = {1},
	pages = {14904},
	issn = {2045-2322},
	doi = {10.1038/s41598-020-71942-7},
	url = {https://www.nature.com/articles/s41598-020-71942-7},
	urldate = {2023-07-24},
	langid = {english},
}

@article{silva-rodriguez_going_2020,
	title = {Going Deeper through the {{Gleason}} Scoring Scale: {{An}}
	         Automatic End-to-End System for Histology Prostate Grading and
	         Cribriform Pattern Detection},
	shorttitle = {Going Deeper through the {{Gleason}} Scoring Scale},
	author = {Silva-Rodríguez, Julio and Colomer, Adrián and Sales, María A. and
	          Molina, Rafael and Naranjo, Valery},
	date = {2020-10},
	journaltitle = {Computer Methods and Programs in Biomedicine},
	shortjournal = {Computer Methods and Programs in Biomedicine},
	volume = {195},
	pages = {105637},
	issn = {01692607},
	doi = {10.1016/j.cmpb.2020.105637},
	url = {https://linkinghub.elsevier.com/retrieve/pii/S016926072031470X},
	urldate = {2023-07-24},
	langid = {english},
}

@online{tan_efficientnetv2_2021,
	title = {{{EfficientNetV2}}: {{Smaller Models}} and {{Faster Training}}},
	shorttitle = {{{EfficientNetV2}}},
	author = {Tan, Mingxing and Le, Quoc V.},
	date = {2021-06-23},
	publisher = {arXiv},
	eprint = {2104.00298},
	eprinttype = {arXiv},
	eprintclass = {cs},
	doi = {10.48550/arXiv.2104.00298},
	url = {http://arxiv.org/abs/2104.00298},
	urldate = {2025-10-15},
	pubstate = {prepublished},
}

@article{egevad_interobserver_2023,
	title = {Interobserver Reproducibility of Cribriform Cancer in Prostate
	         Needle Biopsies and Validation of {{International Society}} of {{
	         Urological Pathology}} Criteria},
	author = {Egevad, Lars and Delahunt, Brett and Iczkowski, Kenneth A and Van
	          Der Kwast, Theo and Van Leenders, Geert J L H and Leite, Katia R M
	          and Pan, Chin‐Chen and Samaratunga, Hemamali and Tsuzuki, Toyonori
	          and Mulliqi, Nita and Ji, Xiaoyi and Olsson, Henrik and Valkonen,
	          Masi and Ruusuvuori, Pekka and Eklund, Martin and Kartasalo, Kimmo},
	date = {2023-05},
	journaltitle = {Histopathology},
	shortjournal = {Histopathology},
	volume = {82},
	number = {6},
	pages = {837--845},
	issn = {0309-0167, 1365-2559},
	doi = {10.1111/his.14867},
	url = {https://onlinelibrary.wiley.com/doi/10.1111/his.14867},
	urldate = {2025-02-06},
	langid = {english},
}

@article{russo_oncological_2023,
	title = {Oncological Outcomes of Cribriform Histology Pattern in Prostate
	         Cancer Patients: A Systematic Review and Meta-Analysis},
	shorttitle = {Oncological Outcomes of Cribriform Histology Pattern in
	              Prostate Cancer Patients},
	author = {Russo, Giorgio Ivan and Soeterik, Timo and Puche-Sanz, Ignacio and
	          Broggi, Giuseppe and Lo Giudice, Arturo and De Nunzio, Cosimo and
	          Lombardo, Riccardo and Marra, Giancarlo and Gandaglia, Giorgio and
	          {European Association of Urology Young Academic Urologists}},
	date = {2023-12},
	journaltitle = {Prostate Cancer and Prostatic Diseases},
	shortjournal = {Prostate Cancer Prostatic Dis},
	volume = {26},
	number = {4},
	eprint = {36216967},
	eprinttype = {pmid},
	pages = {646--654},
	issn = {1476-5608},
	doi = {10.1038/s41391-022-00600-y},
	langid = {english},
}

@article{kweldam_cribriform_2015,
	title = {Cribriform Growth Is Highly Predictive for Postoperative Metastasis
	         and Disease-Specific Death in {{Gleason}} Score 7 Prostate Cancer},
	author = {Kweldam, Charlotte F and Wildhagen, Mark F and Steyerberg, Ewout W
	          and Bangma, Chris H and Van Der Kwast, Theodorus H and Van Leenders
	          , Geert Jlh},
	date = {2015-03},
	journaltitle = {Modern Pathology},
	shortjournal = {Modern Pathology},
	volume = {28},
	number = {3},
	pages = {457--464},
	issn = {08933952},
	doi = {10.1038/modpathol.2014.116},
	url = {https://linkinghub.elsevier.com/retrieve/pii/S0893395222019159},
	urldate = {2025-02-26},
	langid = {english},
}

@article{osiecki_presence_2024,
	title = {The Presence of Cribriform Pattern in Prostate Biopsy and Radical
	         Prostatectomy Is Associated with Negative Postoperative Pathological
	         Features},
	author = {Osiecki, Rafal and Kozikowski, Mieszko and Białek, Łukasz and
	          Pyzlak, Michał and Dobruch, Jakub},
	date = {2024},
	journaltitle = {Central European Journal of Urology},
	shortjournal = {Cent European J Urol},
	volume = {77},
	number = {1},
	eprint = {38645812},
	eprinttype = {pmid},
	pages = {22--29},
	issn = {2080-4806},
	doi = {10.5173/ceju.2023.215},
	langid = {english},
	pmcid = {PMC11032032},
}

@article{flammia_cribriform_2020,
	title = {Cribriform Pattern Does Not Have a Significant Impact in {{Gleason
	         Score}} $\geq$7/{{ISUP Grade}} $\geq$2 Prostate Cancers Submitted to
	         Radical Prostatectomy},
	author = {Flammia, Simone and Frisenda, Marco and Maggi, Martina and
	          Magliocca, Fabio Massimo and Ciardi, Antonio and Panebianco,
	          Valeria and De Berardinis, Ettore and Salciccia, Stefano and Di
	          Pierro, Giovanni Battista and Gentilucci, Alessandro and Del
	          Giudice, Francesco and Busetto, Gian Maria and Gallucci, Michele
	          and Sciarra, Alessandro},
	date = {2020-09-18},
	journaltitle = {Medicine},
	volume = {99},
	number = {38},
	pages = {e22156},
	issn = {0025-7974, 1536-5964},
	doi = {10.1097/MD.0000000000022156},
	url = {https://journals.lww.com/10.1097/MD.0000000000022156},
	urldate = {2025-02-26},
	langid = {english},
}

@article{ericson_diagnostic_2020,
	title = {Diagnostic {{Accuracy}} of {{Prostate Biopsy}} for {{Detecting
	         Cribriform Gleason Pattern}} 4 {{Carcinoma}} and {{Intraductal
	         Carcinoma}} in {{Paired Radical Prostatectomy Specimens}}: {{
	         Implications}} for {{Active Surveillance}}},
	shorttitle = {Diagnostic {{Accuracy}} of {{Prostate Biopsy}} for {{Detecting
	              Cribriform Gleason Pattern}} 4 {{Carcinoma}} and {{Intraductal
	              Carcinoma}} in {{Paired Radical Prostatectomy Specimens}}},
	author = {Ericson, Kyle J. and Wu, Shannon S. and Lundy, Scott D. and Thomas
	          , Lewis J. and Klein, Eric A. and McKenney, Jesse K.},
	date = {2020-02},
	journaltitle = {Journal of Urology},
	shortjournal = {Journal of Urology},
	volume = {203},
	number = {2},
	pages = {311--319},
	issn = {0022-5347, 1527-3792},
	doi = {10.1097/JU.0000000000000526},
	url = {http://www.auajournals.org/doi/10.1097/JU.0000000000000526},
	urldate = {2025-02-26},
	langid = {english},
}

@article{markl_number_2021,
	title = {Number of Pathologists in {{Germany}}: Comparison with {{European}}
	         Countries, {{USA}}, and {{Canada}}},
	shorttitle = {Number of Pathologists in {{Germany}}},
	author = {Märkl, Bruno and Füzesi, Laszló and Huss, Ralf and Bauer, Svenja
	          and Schaller, Tina},
	date = {2021-02},
	journaltitle = {Virchows Archiv: An International Journal of Pathology},
	shortjournal = {Virchows Arch},
	volume = {478},
	number = {2},
	eprint = {32719890},
	eprinttype = {pmid},
	pages = {335--341},
	issn = {1432-2307},
	doi = {10.1007/s00428-020-02894-6},
	langid = {english},
	pmcid = {PMC7969551},
}

@article{rabilloud_deep_2023,
	title = {Deep {{Learning Methodologies Applied}} to {{Digital Pathology}} in
	         {{Prostate Cancer}}: {{A Systematic Review}}},
	shorttitle = {Deep {{Learning Methodologies Applied}} to {{Digital Pathology
	              }} in {{Prostate Cancer}}},
	author = {Rabilloud, Noémie and Allaume, Pierre and Acosta, Oscar and De
	          Crevoisier, Renaud and Bourgade, Raphael and Loussouarn, Delphine
	          and Rioux-Leclercq, Nathalie and Khene, Zine-eddine and Mathieu,
	          Romain and Bensalah, Karim and Pecot, Thierry and Kammerer-Jacquet,
	          Solene-Florence},
	date = {2023-08-14},
	journaltitle = {Diagnostics},
	shortjournal = {Diagnostics},
	volume = {13},
	number = {16},
	pages = {2676},
	issn = {2075-4418},
	doi = {10.3390/diagnostics13162676},
	url = {https://www.mdpi.com/2075-4418/13/16/2676},
	urldate = {2025-02-26},
	langid = {english},
}

@article{gronberg_prostate_2015,
	title = {Prostate Cancer Screening in Men Aged 50–69 Years ({{STHLM3}}): A
	         Prospective Population-Based Diagnostic Study},
	shorttitle = {Prostate Cancer Screening in Men Aged 50–69 Years ({{STHLM3}})
	              },
	author = {Grönberg, Henrik and Adolfsson, Jan and Aly, Markus and Nordström,
	          Tobias and Wiklund, Peter and Brandberg, Yvonne and Thompson, James
	          and Wiklund, Fredrik and Lindberg, Johan and Clements, Mark and
	          Egevad, Lars and Eklund, Martin},
	date = {2015-12},
	journaltitle = {The Lancet Oncology},
	shortjournal = {The Lancet Oncology},
	volume = {16},
	number = {16},
	pages = {1667--1676},
	issn = {14702045},
	doi = {10.1016/S1470-2045(15)00361-7},
	url = {https://linkinghub.elsevier.com/retrieve/pii/S1470204515003617},
	urldate = {2025-02-27},
	langid = {english},
}

@article{epstein_2014_2016,
	title = {The 2014 {{International Society}} of {{Urological Pathology}} ({{
	         ISUP}}) {{Consensus Conference}} on {{Gleason Grading}} of {{
	         Prostatic Carcinoma}}: {{Definition}} of {{Grading Patterns}} and {{
	         Proposal}} for a {{New Grading System}}},
	shorttitle = {The 2014 {{International Society}} of {{Urological Pathology}}
	              ({{ISUP}}) {{Consensus Conference}} on {{Gleason Grading}} of {
	              {Prostatic Carcinoma}}},
	author = {Epstein, Jonathan I. and Egevad, Lars and Amin, Mahul B. and
	          Delahunt, Brett and Srigley, John R. and Humphrey, Peter A.},
	date = {2016-02},
	journaltitle = {American Journal of Surgical Pathology},
	volume = {40},
	number = {2},
	pages = {244--252},
	issn = {0147-5185},
	doi = {10.1097/PAS.0000000000000530},
	url = {https://journals.lww.com/00000478-201602000-00010},
	urldate = {2025-04-03},
	langid = {english},
}

@online{mulliqi_foundation_2025,
	title = {Foundation {{Models}} -- {{A Panacea}} for {{Artificial
	         Intelligence}} in {{Pathology}}?},
	author = {Mulliqi, Nita and Blilie, Anders and Ji, Xiaoyi and Szolnoky,
	          Kelvin and Olsson, Henrik and Boman, Sol Erika and Titus, Matteo
	          and Gonzalez, Geraldine Martinez and Mielcarz, Julia Anna and
	          Valkonen, Masi and Gudlaugsson, Einar and Kjosavik, Svein R. and
	          Asenjo, José and Gambacorta, Marcello and Libretti, Paolo and Braun
	          , Marcin and Kordek, Radzislaw and Łowicki, Roman and Hotakainen,
	          Kristina and Väre, Päivi and Pedersen, Bodil Ginnerup and Sørensen,
	          Karina Dalsgaard and Ulhøi, Benedicte Parm and Ruusuvuori, Pekka
	          and Delahunt, Brett and Samaratunga, Hemamali and Tsuzuki, Toyonori
	          and Janssen, Emilius A. M. and Egevad, Lars and Eklund, Martin and
	          Kartasalo, Kimmo},
	date = {2025-03-03},
	publisher = {arXiv},
	eprint = {2502.21264},
	eprinttype = {arXiv},
	eprintclass = {cs},
	doi = {10.48550/arXiv.2502.21264},
	url = {http://arxiv.org/abs/2502.21264},
	urldate = {2025-04-03},
	pubstate = {prepublished},
}

@article{mulliqi_development_2025,
	title = {Development and Retrospective Validation of an Artificial
	         Intelligence System for Diagnostic Assessment of Prostate Biopsies:
	         Study Protocol},
	shorttitle = {Development and Retrospective Validation of an Artificial
	              Intelligence System for Diagnostic Assessment of Prostate
	              Biopsies},
	author = {Mulliqi, Nita and Blilie, Anders and Ji, Xiaoyi and Szolnoky,
	          Kelvin and Olsson, Henrik and Titus, Matteo and Martinez Gonzalez,
	          Geraldine and Boman, Sol Erika and Valkonen, Masi and Gudlaugsson,
	          Einar and Kjosavik, Svein Reidar and Asenjo, José and Gambacorta,
	          Marcello and Libretti, Paolo and Braun, Marcin and Kordek,
	          Radzislaw and Łowicki, Roman and Hotakainen, Kristina and Väre,
	          Päivi and Pedersen, Bodil Ginnerup and Sørensen, Karina Dalsgaard
	          and Ulhøi, Benedicte Parm and Rantalainen, Mattias and Ruusuvuori,
	          Pekka and Delahunt, Brett and Samaratunga, Hemamali and Tsuzuki,
	          Toyonori and Janssen, Emilius Adrianus Maria and Egevad, Lars and
	          Kartasalo, Kimmo and Eklund, Martin},
	date = {2025-07},
	journaltitle = {BMJ Open},
	shortjournal = {BMJ Open},
	volume = {15},
	number = {7},
	pages = {e097591},
	issn = {2044-6055, 2044-6055},
	doi = {10.1136/bmjopen-2024-097591},
	url = {https://bmjopen.bmj.com/lookup/doi/10.1136/bmjopen-2024-097591},
	urldate = {2025-08-10},
	langid = {english},
}

@online{boman_impact_2025,
	title = {The Impact of Tissue Detection on Diagnostic Artificial
	         Intelligence Algorithms in Digital Pathology},
	author = {Boman, Sol Erika and Mulliqi, Nita and Blilie, Anders and Ji,
	          Xiaoyi and Szolnoky, Kelvin and Gudlaugsson, Einar and Janssen,
	          Emiel A. M. and Kjosavik, Svein R. and Asenjo, José and Gambacorta,
	          Marcello and Libretti, Paolo and Braun, Marcin and Kordek,
	          Radzislaw and Łowicki, Roman and Hotakainen, Kristina and Väre,
	          Päivi and Pedersen, Bodil Ginnerup and Sørensen, Karina Dalsgaard
	          and Ulhøi, Benedicte Parm and Egevad, Lars and Kartasalo, Kimmo},
	date = {2025},
	publisher = {arXiv},
	doi = {10.48550/ARXIV.2503.23021},
	url = {https://arxiv.org/abs/2503.23021},
	urldate = {2025-08-10},
	pubstate = {prepublished},
	version = {1},
}

@article{van_leenders_2019_2020,
	title = {The 2019 {{International Society}} of {{Urological Pathology}} ({{
	         ISUP}}) {{Consensus Conference}} on {{Grading}} of {{Prostatic
	         Carcinoma}}},
	author = {van Leenders, Geert J. L. H. and van der Kwast, Theodorus H. and
	          Grignon, David J. and Evans, Andrew J. and Kristiansen, Glen and
	          Kweldam, Charlotte F. and Litjens, Geert and McKenney, Jesse K. and
	          Melamed, Jonathan and Mottet, Nicholas and Paner, Gladell P. and
	          Samaratunga, Hemamali and Schoots, Ivo G. and Simko, Jeffry P. and
	          Tsuzuki, Toyonori and Varma, Murali and Warren , Anne Y. and
	          Wheeler, Thomas M. and Williamson, Sean R. and Iczkowski, Kenneth
	          A. and {ISUP Grading Workshop Panel Members}},
	date = {2020-08},
	journaltitle = {The American Journal of Surgical Pathology},
	shortjournal = {Am J Surg Pathol},
	volume = {44},
	number = {8},
	eprint = {32459716},
	eprinttype = {pubmed},
	pages = {e87-e99},
	issn = {1532-0979},
	doi = {10.1097/PAS.0000000000001497},
	langid = {english},
	pmcid = {PMC7382533},
}

@article{van_der_kwast_isup_2021,
	title = {{{ISUP Consensus Definition}} of {{Cribriform Pattern Prostate
	         Cancer}}},
	author = {van der Kwast, Theodorus H. and van Leenders, Geert J. and Berney,
	          Daniel M. and Delahunt, Brett and Evans, Andrew J. and Iczkowski,
	          Kenneth A. and McKenney, Jesse K. and Ro, Jae Y. and Samaratunga,
	          Hemamali and Srigley, John R. and Tsuzuki, Toyo and Varma, Murali
	          and Wheeler, Thomas M. and Egevad, Lars},
	date = {2021-08-01},
	journaltitle = {The American Journal of Surgical Pathology},
	shortjournal = {Am J Surg Pathol},
	volume = {45},
	number = {8},
	eprint = {33999555},
	eprinttype = {pubmed},
	pages = {1118--1126},
	issn = {1532-0979},
	doi = {10.1097/PAS.0000000000001728},
	langid = {english},
}
\end{refsection}

\end{document}